\documentclass[runningheads]{llncs}

\usepackage{eccv}

\usepackage{eccvabbrv}

\usepackage{graphicx}
\usepackage{booktabs}
\usepackage{multirow}
\usepackage{placeins} 
\usepackage{tabularx}
\usepackage{array}
\usepackage{booktabs}
\usepackage{amssymb}
\usepackage{tcolorbox}
\tcbuselibrary{breakable}
\usepackage[accsupp]{axessibility}  

\usepackage{hyperref}

\usepackage{orcidlink}

\begin{document}

\title{SSMNBench: Diagnosing Image-based Cross-View Human-Object Understanding via Single-View Sufficiency and Multi-View Necessity} 

\titlerunning{SSMNBench}

\author{Tianchen Guo\inst{1,4} \and
Chen Liu\inst{3} \and
Ling Chen\inst{3} \and
Xin Yu\inst{2,4}\thanks{Corresponding author.}}

\authorrunning{T.~Guo et al.}

\institute{The University of Queensland \\
\email{tianchen.guo@uq.edu.au} \and
Australian Institute for Machine Learning, Adelaide University \and
University of Technology Sydney \and
Follow Me AI Pty LTD}

\maketitle

\begin{abstract}
Multimodal Large Language Models (MLLMs) have shown remarkable progress in single-image perception, yet their ability to reason about complex cross-view human-centric scenes remains largely unverified. Current multi-view benchmarks evaluate models using a fixed ``bag of frames'' and thus conflate a model's robustness to visual distraction with its genuine ability to fuse fragmented cross-view evidence. To address this issue, we introduce SSMNBench, a diagnostic benchmark comprising 3,300 curated QA pairs for cross-view human and human-object understanding. SSMNBench uniquely categorizes tasks into Single-View Sufficiency (SVS) and Multi-View Necessity (MVN). By systematically perturbing view availability across 17 state-of-the-art MLLMs, critical limitations are revealed: models suffer from severe ``distraction degradation'' when presented with redundant views (SVS), and fail to integrate fragmented geometric evidence across cameras (MVN). Our evaluations demonstrate that modern MLLMs rely on multiple single-image semantic averaging and view preference rather than genuine cross-view synthesis. By exposing these fundamental vulnerabilities, SSMNBench provides a rigorous diagnostic framework to drive the advancement of future cross-view-aware multimodal architectures. The code is available at: \url{https://github.com/gtc-gh/SSMNBench}
\end{abstract}

\section{Introduction}
\label{sec:intro}

Multi-modal Large Language Models (MLLMs)~\cite{wu2023multimodal,yin2024survey,yang2025qwen3,chen2024internvl,bai2023qwen,deepseekai2024deepseekv3technicalreport,liu2023llava,wang2025internvl3_5,dubey2024llama,comanici2025gemini,qwen3.5,team2023gemini,openai2025o3o4mini,deepmind2025gemini25} have rapidly advanced visual understanding~\cite{cho2025perceptionlm,shen2025vlm,chen2025r1v,openr1,sun2025reinforcementfinetuningpowersreasoning,vteam2025glm45vglm41vthinkingversatilemultimodal, guo2026beyond,qi2026smokebench,wu2026metom,wu2022showface,zhang2025mllms,xu2024m3a,xu2025mdam3}, yet most evaluation benchmarks still reflect a \textit{single-view} scenario: a model sees one single view and answers one question~\cite{fu2023mme,zhang2025tiu,huang2025ocr,nguyen2025localizing,qin2025face,zhou2025robotracer,anywhere3d,ma2025spatialllm,liu2025visual,chang2025wearvqa,jia2025omnispatial}. In contrast, human-centric scenes, where people interact with objects, with diverse poses and frequent self-/object-occlusion, often require \textit{multiple} viewpoints to answer what is happening. This makes multi-view understanding particularly important for tasks aligned with human observation~\cite{grauman2024ego,xu2025m3gym,ozsoy20224d,ozsoy2024mmor,khirodkar2024harmony4d,zhang2024hoi}, such as activity understanding, interaction recognition, and human attribute/role reasoning, where crucial evidence may be hidden from one camera but revealed by another.

Despite growing interest in multi-view benchmarks~\cite{yeh2025seeing,yang2025mmsi,fu2024blink,wang2024muirbench}, a key methodological gap remains: most protocols simply provide a ``bag of frames'' (a fixed set of views) and report final accuracy. This conflates two distinct capabilities. First, a model may only need \textit{one} view, but must remain robust when extra (redundant) views are present. Second, a model may truly need to \textit{fuse} complementary evidence across views when no single view suffices. Without separating these cases, accuracy alone cannot diagnose whether failures come from distraction by redundancy or inability to integrate fragmented cross-view cues.

We address this gap by introducing \textbf{SSMNBench}, a human-centred benchmark for \textit{cross-view human and human--object understanding}. SSMNBench contains \textbf{11 tasks} and \textbf{3300} manually curated question--answer (QA) pairs (\textbf{300} QA per task) sourced from diverse data~\cite{xu2025m3gym,ozsoy20224d,ozsoy2024mmor,gan2021mvmhat,khirodkar2024harmony4d,zhang2024hoi,khirodkar2023ego,liu2025core4d} and then carefully selected through manual review. Questions cover both humans, human-related objects, and human-object interaction, where occlusion, viewpoint-dependent appearance, and pose diversity further impose challenges on model reasoning.

Central to SSMNBench is an evaluation taxonomy that distinguishes \textbf{Single-View Sufficiency (SVS)} from \textbf{Multi-View Necessity (MVN)} based on whether the required evidence is contained within a single view or distributed across multiple views:
\begin{itemize}
    \item \textbf{Single-View Sufficiency (SVS).} There exists at least one view in which the question is answerable on its own. Concretely, we annotate a ``Golden View'' $V_{GT}$ from which the answer can be derived; importantly, \emph{other} views may also be sufficient (SVS does not imply uniqueness). SVS therefore tests whether MLLMs can identify and use an informative view \emph{without being distracted} by additional redundant viewpoints.
    \item \textbf{Multi-View Necessity (MVN).} No single view contains enough evidence to answer the question reliably; the model must combine complementary cues across views. MVN therefore tests whether MLLMs can perform \emph{cross-view integration} under occlusion and viewpoint-dependent ambiguity, rather than relying on semantic priors.
\end{itemize}

To probe these capabilities systematically, we evaluate both SVS and MVN under \textbf{five} view-availability settings: \textit{Normal}, \textit{+1}, \textit{+2}, \textit{+3} (increasing numbers of additional views), and \textit{-1} (removing one view). These controlled perturbations reveal not only whether a model answers correctly, but also how its performance \textit{changes} as views are added or removed. Accordingly, besides accuracy, we recommend reporting a \textit{Distraction Decay $\delta_{dis}$} metric that summarizes performance variation across settings (\ie, the accuracy drop from \textit{Normal} to \textit{+k} for SVS), capturing distraction and reliance on missing evidence.

Using this protocol, we benchmark \textbf{17} MLLMs and find that multi-view input is not automatically beneficial: additional views can actively \textit{hurt} SVS due to redundancy-induced distraction, while MVN exposes brittle geometric fusion and increased hallucination when critical evidence is fragmented. Ultimately, SSMNBench provides a targeted diagnostic platform for developing MLLMs capable of true cross-view understanding in complex, real-world multi-view environments.

In summary, our main contributions are:
\begin{itemize}
    \item \textbf{Novel Evaluation Taxonomy:} We introduce the first diagnostic framework that distinguishes \textbf{Single-View Sufficiency (SVS)} from \textbf{Multi-View Necessity (MVN)} for MLLMs, explicitly separating a model's robustness to visual distraction from its capacity for cross-view fusion.
    \item \textbf{Comprehensive Benchmark:} We present SSMNBench, containing 3,300 rigorously curated, expert-annotated QA pairs spanning 11 diverse tasks focused on dense, occlusion-heavy, cross-view human-centric understanding.
    \item \textbf{Systematic Diagnostic Findings:} Through controlled view perturbation and the proposed Distraction Decay ($\delta_{dis}$) metric, we reveal that current architectures universally suffer from context saturation and over-rely on monocular priors rather than performing genuine 3D cross-view synthesis.
\end{itemize}

\section{Related Work}
\label{sec:related_work}

\subsection{Multi-View and Multi-Image Benchmarks for MLLMs}
Multimodal tasks increasingly move beyond isolated perception and require models to jointly interpret multiple inputs \cite{liu2024compound, liu2024affective, zhang2024effective, liu2024benchmarking, qiu2024language, liu2025robust, liu2025dynamic, zhang2024affective, guo2024being, qiu2024learning}, and associate entities across modalities.
As MLLMs~\cite{bai2023qwen,chen2024internvl,yang2025qwen3,yin2024survey,deepseekai2024deepseekv3technicalreport,liu2023llava,lu2024ovis,touvron2023llama,pan2024large} evolve beyond single-image comprehension, a growing body of literature has sought to evaluate their reasoning across multiple visual inputs. Recent benchmarks~\cite{ke2025dynamic,qi2026smokebench,guo2026beyond,hong2026esi,fu2026mme,guo2025plnet} such as VSI-Bench~\cite{yang2025thinking}, AllAnglesBench~\cite{yeh2025seeing}, and Ego3DBench~\cite{gholami2025spatialreasoningvisionlanguagemodels} test geometric awareness under camera changes, while benchmarks like MuirBench~\cite{wang2024muirbench}, Blink~\cite{fu2024blink}, and MMSI-Bench~\cite{yang2025mmsi} require models to link shared entities and attributes across disparate images. These works consistently highlight that current architectures struggle with cross-image correspondence and visual aggregation.

However, a fundamental methodological limitation persists: evaluations treat inputs as a fixed ``bag of frames'' and report a single end-to-end accuracy metric~\cite{yeh2025seeing, li2024mvbench, guo2026beyond}. 
Consequently, this scoring method inadvertently rewards the exploitation of visual and linguistic biases, generating metrics that overestimate actual capabilities since models rely on single unobstructed frames rather than performing genuine 3D spatial integration~\cite{yeh2025seeing,yang2025mmsi,gholami2025spatialreasoningvisionlanguagemodels}.
This paradigm conflates two distinct cognitive capabilities. It cannot determine whether a model succeeds through genuine cross-view fusion or simply by attending to a single informative view. Conversely, it cannot determine whether failures stem from redundancy-induced distraction or an inability to integrate fragmented evidence under occlusion. SSMNBench directly addresses this blind spot. By categorizing tasks into SVS and MVN and systematically perturbing view availability, we explicitly decouple genuine spatial integration from reasoning driven by semantic priors.

\vspace{-1em}
\subsection{Human-Centric Understanding in Crowded 3D Scenes}
\vspace{-0.5em}
Human-centric visual understanding has advanced through single-subject datasets featuring clean, unambiguous viewpoints, such as Ego-Exo4D~\cite{grauman2024ego} and H3WB~\cite{Zhu_2023_ICCV}. However, they lack sufficient representation of the identity ambiguity, mutual occlusion, and viewpoint dependency that characterize unstructured real-world environments~\cite{khirodkar2023ego,liu2025core4d}. In these dynamic settings, bodies frequently overlap, and critical visual evidence is often partially or entirely hidden from the single camera. Cross-view reasoning is therefore necessary when single views cannot provide sufficient information. 
While recent explicit 3D reconstruction and rendering techniques, such as 3D Gaussian Splatting~\cite{du20253drealcar,du2024mvgs,du2026mobile,du2024dreamcar,du2024ethics,chen2024survey}, have revolutionized high-fidelity scene and human-centric representation, modern MLLMs still learn to implicitly fuse geometric cues directly from sparse 2D frames.
Recent multi-view datasets capture occlusion-dense, multi-person scenes~\cite{xu2025m3gym,ozsoy20224d,ozsoy2024mmor,gan2021mvmhat} and complex interactions~\cite{khirodkar2024harmony4d,zhang2024hoi}, where distinguishing physical contact details from near-contact and reasoning complex spatial relationships often require triangulating evidence from specific and complementary angles. 

While these datasets predominantly target low-level geometric tasks, such as 3D tracking or pose estimation, and offer valuable raw multi-camera video, they lack the high-level semantic question-answering annotations required to benchmark the ability of cross-view reasoning and cognition of modern MLLMs for diverse tasks. 
SSMNBench bridges this gap by building a rigorous semantic evaluation on these occlusion-heavy sources, creating a platform where genuine cross-view integration is a strict prerequisite for correctly answering questions.

\begin{figure}[tb]
  \centering
  \includegraphics[height=6.5cm]{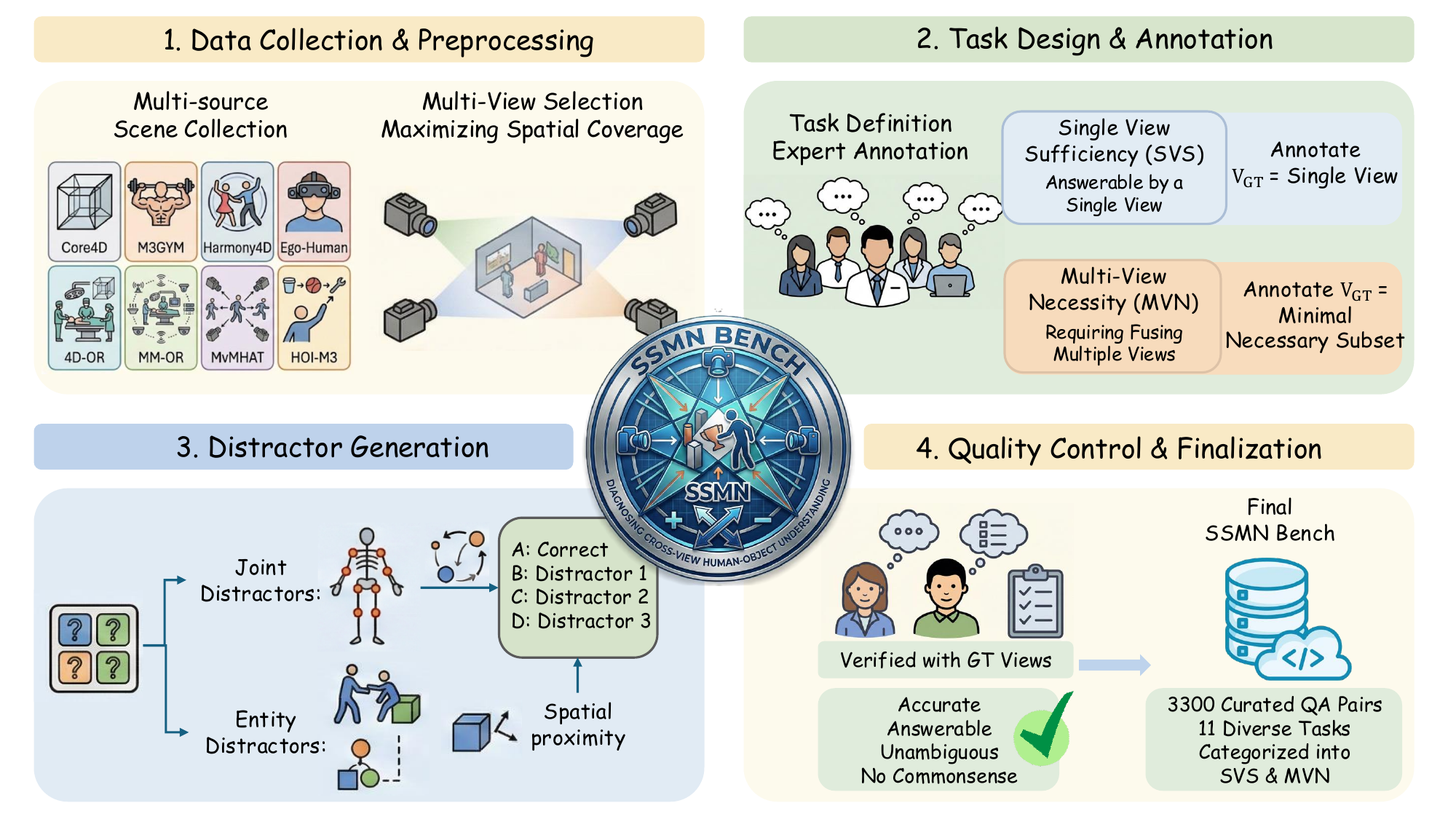}
  \vspace{-0.5em}
  \caption{Illustration of the SSMNBench curation pipeline. The construction process begins by collecting dense, occlusion-heavy multi-view scenes and defining 11 distinct SVS and MVN tasks. Next, experts annotate QA pairs alongside their necessary ground-truth views. Finally, we generate structured distractors to mitigate linguistic priors, enforce strict quality control via blind verification, and randomize input orders to eliminate camera positional bias.}
  \label{fig:main}
  \vspace{-1.0em}
\end{figure}

\section{Proposed SSMNBench Framework}
\label{sec:framework}
\subsection{Overview}
Current multi-view benchmarks typically aggregate all available visual information, obscuring the specific contribution of individual viewpoints. \textbf{SSMNBench} departs from this paradigm by explicitly modeling the epistemic relationship between visual input and semantic understanding. Our framework distinguishes \textbf{SVS} from \textbf{MVN} to diagnose whether multi-modal large language models (MLLMs) fail due to redundancy-induced distraction or an inability to integrate fragmented cross-view cues. 

\vspace{-1em}
\subsection{Benchmark Construction Process}
\vspace{-0.5em}
To ensure a rigorous evaluation of spatial reasoning across diverse scenarios, we develop a comprehensive 6-step benchmark construction pipeline. This encompasses data collection and preprocessing, meticulous task design, multi-stage expert question-answering annotation, structured distractor generation, strict quality control, and post-processing to eliminate dataset biases. The overall pipeline is illustrated in Figure~\ref{fig:main}.

\vspace{0.5em}
\noindent\textbf{3.2.1 Data Collection and Preprocessing}
\vspace{-1em}

\paragraph{Data Collection.} To evaluate complex cross-view understanding, we skip simple, isolated actions and explicitly target scenarios characterized by dense human-centric scenes, multi-person interactions, and high levels of mutual- and self-occlusion. We manually collect and curate data from diverse, representative real-world multi-view datasets (\ie, Core4D~\cite{liu2025core4d}, M3GYM~\cite{xu2025m3gym}, Harmony4D~\cite{khirodkar2024harmony4d}, Ego-Human~\cite{khirodkar2023ego}, 4D-OR~\cite{ozsoy20224d}, MM-OR~\cite{ozsoy2024mmor}, MvMHAT~\cite{gan2021mvmhat}, and HOI-M3~\cite{zhang2024hoi}). This deliberate focus ensures our benchmark reflects the inherent complexity and occlusion density of unstructured real-world environments. 

\vspace{-1em}
\paragraph{Preprocessing.} For each selected scene, we synchronize multi-view videos and extract frame pairs. 
To construct the multi-view input, we select exactly four camera views that exhibit minimal field-of-view (FoV) overlap while collectively maximizing the spatial coverage and information completeness of scenes.

\vspace{0.5em}
\noindent\textbf{3.2.2 Task Definition and Design}
\vspace{0.5em}

\begin{figure}[tb]
  \centering
  \includegraphics[height=6.5cm]{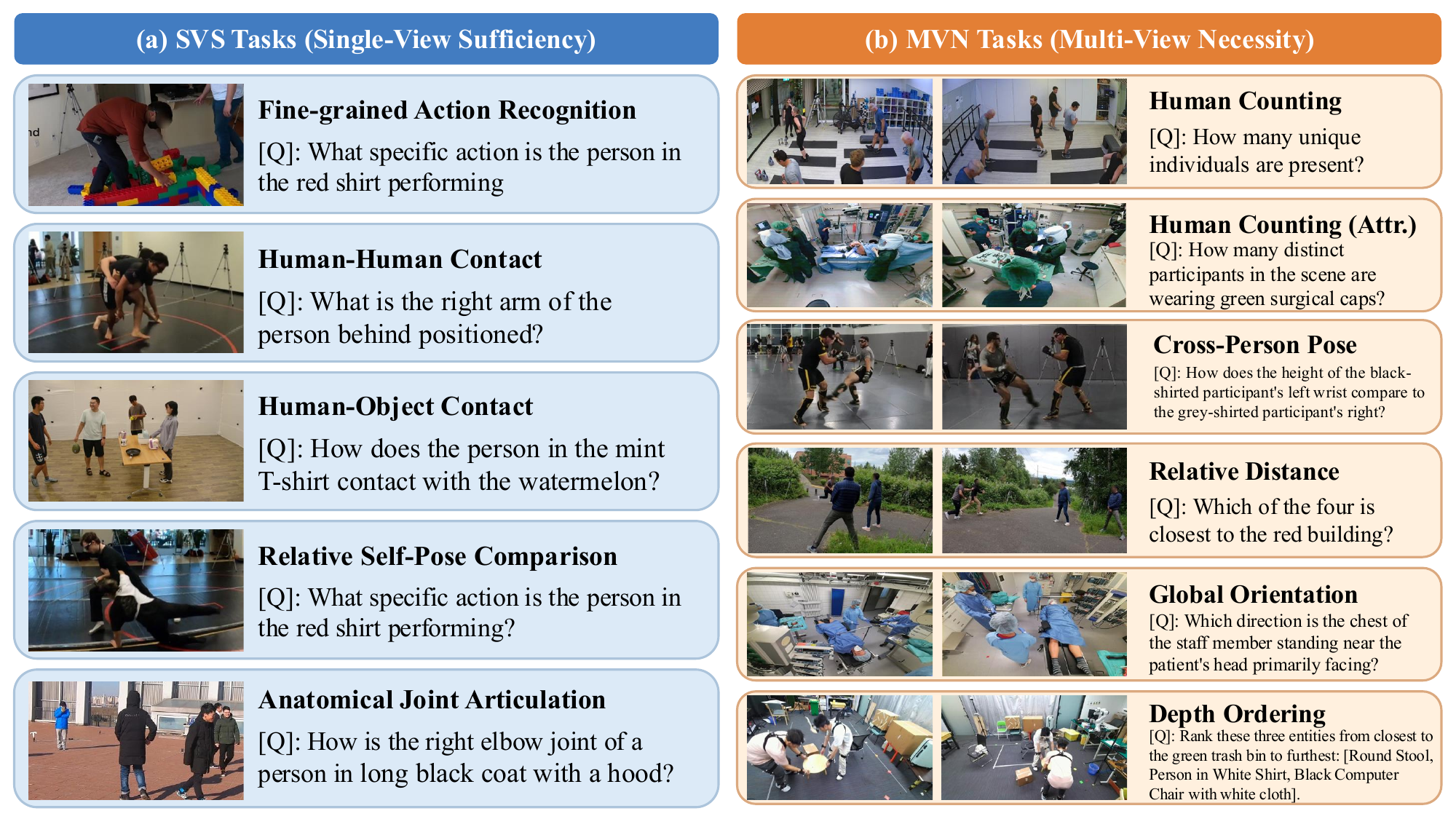}
  \vspace{-0.5em}
  \caption{Visual examples of the 11 tasks in SSMNBench, categorized by their reliance on view sufficiency. SVS tasks can be resolved with a single clear view, while MVN tasks require synthesizing information from multiple viewpoints to overcome occlusion and ambiguity.}
  \label{fig:task_demo}
\end{figure}

\begin{figure}[tb]
  \centering
  \includegraphics[height=6.5cm]{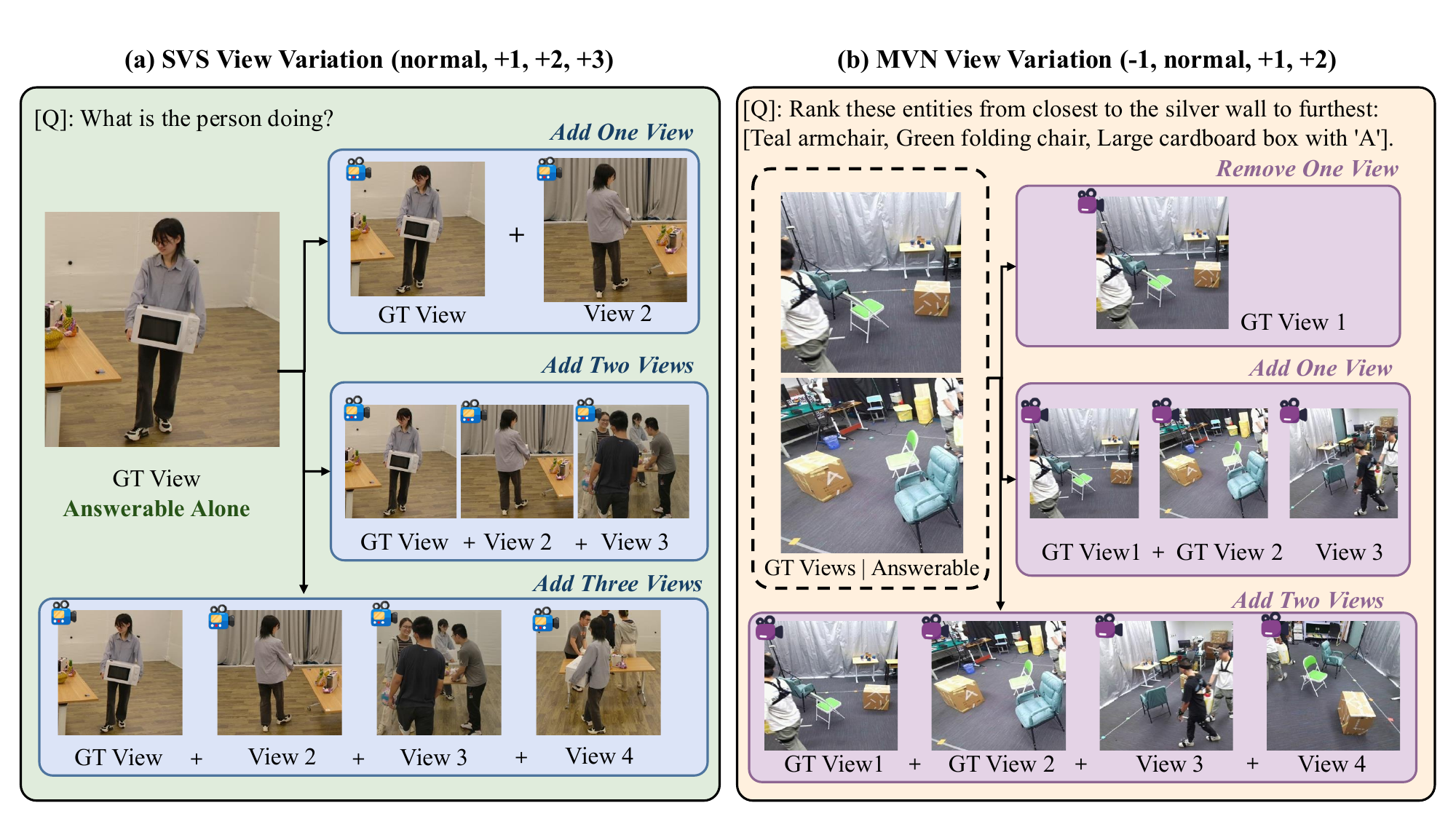}
  \vspace{-0.5em}
  \caption{Illustration of view variation in SSMNBench's SVS and MVN settings.}
  \label{fig:view_variation}
\end{figure}

\noindent We define 11 distinct tasks to probe specific aspects of human-centric understanding, categorized by their reliance on visual sufficiency versus necessity. Details and visual examples for each task are provided in Figure~\ref{fig:task_demo} and Figure~\ref{fig:view_variation}.

\vspace{-1em}
\paragraph{SVS Tasks: Fine-Grained Perception.} These tasks test the model's ability to extract subtle visual details when a single sufficient viewpoint is available:
\vspace{-0.5em}
\begin{itemize}
    \item \textbf{Fine-grained Action Recognition} -- Recognizing specific motions and distinguishing between subtle, semantically similar physical actions or states from a clear viewpoint.
    \item \textbf{Human-Human Contact} -- Identifying specific physical contact points between individuals and distinguishing actual contact from near-misses.
    \item \textbf{Human-Object Contact} -- Identifying precise physical interactions between individuals and objects, distinguishing actual contact from near-misses, and localizing the specific contact area.
    \item \textbf{Relative Self-Pose Comparison} -- Comparing the height and relative positioning of one anatomical joint with another on the same individual.
    \item \textbf{Anatomical Joint Articulation} -- Determining whether a specific joint is in flexion or extension, and estimating the degree of flexion (\eg, obtuse/slightly flexed, right angle, or acute/deeply flexed).
\end{itemize}

\vspace{-1em}
\paragraph{MVN Tasks: Global Cross-View Reasoning.} These tasks require building coherent cross-view associations, as no single view provides the complete information for answering. Cross-view integration is required here to resolve depth ambiguities, overcome severe occlusions, and integrate complementary object and human visibilities:
\vspace{-0.5em}
\begin{itemize}
    \item \textbf{Distinct Human Counting} -- Counting unique human entities in crowded scenes where individuals may be fully or partially occluded in certain views, or outside the field of view in others.
    \item \textbf{Distinct Human Counting with Attribute} -- Counting unique entities that match specific semantic or behavioral criteria across different views.
    \item \textbf{Relative Cross-Person Pose Comparison} -- Evaluating 3D posture differences between multiple distinct individuals whose full bodies cannot be captured simultaneously by any single camera.
    \item \textbf{Relative Distance Estimation} -- Measuring the 3D spatial gap between multiple entities (human-human or human-object) to determine proximity (\eg, identifying which entity is closer to a target). This requires cross-view triangulation to resolve monocular depth ambiguity.
    \item \textbf{Global Orientation Identification} -- Identifying the global directional facing of a human's torso or head by linking visual cues across multiple viewpoints (\eg, determining which object or direction a person is currently looking toward).
    \item \textbf{Global Depth Ordering} -- Ranking multiple entities based on their relative distance from a reference point, necessitating a unified spatial understanding to eliminate ambiguous depth projections inherent in single viewpoints.
\end{itemize}

\vspace{0.5em}
\noindent\textbf{3.2.3 Question-Answer Design and Annotation}
\vspace{0.5em}

\noindent To provide rigorous and physically grounded ground truth, we assemble a team of 6 researchers specializing in computer vision and biomechanics. Annotators are asked to create diverse, challenging question-answer (QA) pairs that remain factually consistent regardless of the presence of additional views. 

For the view annotation process, annotators manually examine the image sets to determine the Ground-Truth View Set $\mathcal{V}_{GT}$:
\begin{itemize}
    \item \textbf{For SVS:} Annotators manually identify a single view from which the question is definitively answerable by a human. If multiple views were individually sufficient, the annotator randomly selected one clearly informative view to serve as the golden view.
    \item \textbf{For MVN:} Annotators need to select the \textit{minimal necessary subset} of views (two or more). They also verify that these combined views provide necessary complementary information, which can resolve monocular depth or distance ambiguity as well as eliminate incorrect options.
\end{itemize}
To eliminate individual subjective bias in view selection, each question type is annotated by at least three independent annotators, and the final dataset represents a consensus-driven mixture of annotations.

\vspace{0.5em}
\noindent\textbf{3.2.4 Distractor Generation}
\vspace{0.5em}

\noindent We formulate SSMNBench as a multiple-choice benchmark, where each question is accompanied by four options containing only one uniquely correct answer. To prevent models from exploiting linguistic priors, we employ a structured distractor generation strategy:
\begin{itemize}
    \item \textbf{Human Joint Distractors:} For pose and joint-related questions, distractors are chosen from anatomically adjacent or visually similar joints. Following COCO-style definitions, our benchmark utilizes a comprehensive joint taxonomy: top of head, face, chin, shoulder, chest, upper arm, forearm, wrist, hands, fingers, hips, waist, thighs, lower legs, foot, heel, and toes.
    \item \textbf{Object/Entity Distractors:} For object-related or spatial questions, we deterministically generate three distractors: one spatially closest to the target, one most similar in appearance to the target, and a randomly selected entity from the scene.
\end{itemize}

\noindent\textbf{3.2.5 Quality Control}
\vspace{0.5em}

\noindent We conduct blind verification to ensure QA quality. Reviewers are presented with only the text question and the annotated $\mathcal{V}_{GT}$ (without the remaining views). Any questions that could not be confidently and clearly answered by human reviewers using only $\mathcal{V}_{GT}$ have been removed or modified. Furthermore, we actively filter out ``commonsense'' and ``text-solvable'' questions for which an LLM backbone can guess correct answers purely from linguistic priors without relying on visual evidence.

\vspace{0.5em}
\noindent\textbf{3.2.6 Post-Processing}
\vspace{0.5em}

\noindent During preliminary testing, we observed that MLLMs often exhibit a ``positional bias''~\cite{tian2025identifying,chaudhary2025investigating} based on the order in which images are provided in the prompt. To mitigate input camera positional bias, we apply a uniform randomization strategy to the view ordering. For the raw images sourced from physical cameras 1 through 4, we randomly sample and assign them such that each logical input prompt position (View 1, View 2, View 3, View 4) consists of an approximately 25\% uniform distribution of the original physical camera angles. 

\begin{table}[tb]
  \centering
  \begin{minipage}[t]{0.63\textwidth}
    \centering
    \caption{Task distribution and ground-truth view requirements. $|\mathcal{V}_{GT}|$ denotes the average number of necessary views.}
    \label{tab:task_dist}
    \vspace{-0.5em}
    \resizebox{\textwidth}{!}{
    \begin{tabular}{@{}llcc@{}}
      \toprule
      \textbf{Cat.} & \textbf{Task Name} & \textbf{QA Pairs} & \textbf{Avg.} $|\mathcal{V}_{GT}|$ \\
      \midrule
      \multirow{5}{*}{SVS} 
      & Fine-grained Action & 300 & 1.00 \\
      & Human-Human Contact & 300 & 1.00 \\
      & Human-Object Contact & 300 & 1.00 \\
      & Relative Self-Pose & 300 & 1.00 \\
      & Anatomical Joint & 300 & 1.00 \\
      \midrule
      \multirow{6}{*}{MVN} 
      & Distinct Human Counting & 300 & 2.08 \\
      & Dist. Human Count w/ Attr. & 300 & 2.14 \\
      & Relative Cross-Person Pose & 300 & 2.01 \\
      & Relative Distance Est. & 300 & 2.03 \\
      & Global Orientation Ident. & 300 & 2.06 \\
      & Global Depth Ordering & 300 & 2.06 \\
      \midrule
      \textbf{Total} & & \textbf{3300} & \\
      \bottomrule
    \end{tabular}
    }
  \end{minipage}\hfill
  \begin{minipage}[t]{0.34\textwidth}
    \centering
    \caption{Question-length and answer distributions.}
    \label{tab:dataset_stats}
        \vspace{-0.5em}
    \resizebox{\textwidth}{!}{
    \begin{tabular}{@{}lc@{}}
      \toprule
      \textbf{Metric} & \textbf{Value} \\
      \midrule
      \textbf{Text Length Statistics} & \\
      Avg. Question Length & 117.5 \\
      Avg. Option Length & 42.1 \\
      Avg. Text Option Length & 51.1 \\
      Avg. GT Answer Length & 41.3 \\
      Max Question Length & 319 \\
      Max GT Answer Length & 185 \\
      \midrule
      \textbf{MVN View Distribution} & \\
      2 Views & 93.7\% \\
      3 Views & 6.3\% \\
      \midrule
      \textbf{Answer Distribution} & \\
      Option A & 825 \\
      Option B & 825 \\
      Option C & 825 \\
      Option D & 825 \\
      \bottomrule
    \end{tabular}
    }
  \end{minipage}
\end{table}

\vspace{-1.5em}
\subsection{Dataset Statistics}
The finalized SSMNBench comprises approximately 3,300 high-quality QA pairs evenly distributed across the eleven tasks. Detailed statistics, including task category distributions, correct answer option distributions (A/B/C/D), and text lengths, are summarized in Table~\ref{tab:task_dist} and Table~\ref{tab:dataset_stats}. 


\vspace{-0.5em}
\subsection{Evaluation Metric: Distraction Decay ($\delta_{dis}$)}
\label{sec:metric_distraction}
\vspace{-0.5em}
Standard multi-view benchmarks typically only report an accuracy metric, which masks a model's vulnerability to context saturation. To quantify robustness against redundant visual information, we introduce \textbf{Distraction Decay ($\delta_{dis}$)}. 

This metric measures the average performance drop when additional informative and uninformative views are added to the ground-truth set ($\mathcal{V}_{GT}$). It assesses a model's capacity for \textbf{selective visual attention}---the ability to actively focus on necessary visual evidence while suppressing irrelevant noise.

Since SVS and MVN accommodate different maximum numbers of additional views ($+3$ and $+2$, respectively), we compute the decay independently for each subset $T \in \{SVS, MVN\}$ and report their macro-average. Let $Acc(+k)^T$ be the average accuracy on subset $T$ with $k$ redundant views. We define the overall distraction decay as:
\begin{equation}
\begin{aligned}
    \delta_{dis} = \frac{1}{2} \Bigg[ &\Bigg( Acc(+0)^{SVS} - \frac{1}{3}\sum_{k=1}^{3} Acc(+k)^{SVS} \Bigg) \\
    + &\Bigg( Acc(+0)^{MVN} - \frac{1}{2}\sum_{k=1}^{2} Acc(+k)^{MVN} \Bigg) \Bigg].
\end{aligned}
\end{equation}
A lower $\delta_{dis}$ indicates highly effective selective attention, meaning the model successfully ignores distractor views. Conversely, a higher $\delta_{dis}$ reveals that the model's cross-attention is hijacked by redundant views, diluting the visual signal and degrading performance.

\section{Experiments and Main Findings}
\label{sec:experiments}
\subsection{Experimental Setup}
\paragraph{Evaluated Models.} 
To provide a comprehensive and rigorous assessment of the current MLLMs, we benchmark a diverse suite of 17 state-of-the-art models, including both proprietary models~\cite{openai2025o3o4mini,deepmind2025gemini25,comanici2025gemini} and open-source models~\cite{qwen3.5,wang2025internvl3_5,an2025llava,wu2024deepseek,Qwen2.5-VL}. All images are resized to 1920 $\times$ 1080 to ensure adequate clarity and detail. The main results are shown in Table~\ref{tab:main_results_part1}, and the performance of the other models is provided in the Appendix due to the page limits.

\vspace{-0.5em}
\paragraph{Baselines.}
We report three baselines: (i) \textbf{Random Guess}, corresponding to the accuracy expected from uniformly sampling one option; (ii) \textbf{Blind (Text-Only)}, implemented by evaluating Gemini-2.5-Flash on prompts and options while omitting all visual inputs, to quantify exploitable linguistic bias; and (iii) a \textbf{Human Performance} computed as the average accuracy of five independent graduate-level evaluators (computer science/biomechanics) who were not involved in annotation, under the \textit{Normal ($\mathcal{V}_{GT}$), -1, and +2} settings.

\vspace{-0.5em}
\paragraph{Evaluation Metrics.}
Accuracy (\%) and Distraction Decay $\delta_{dis}$ (\%) are reported for all the experiments. Accuracy is calculated by using an exact match by applying strict regular expressions to extract explicitly formatted final answers. If the rule-based parser fails due to format non-compliance, we apply the LLM-based (\ie Gemini-2.5-Flash-Lite) fallback strategy, read the model's raw text output, and extract the intended choice.


\begin{table}
\centering
\caption{Comprehensive evaluation results on SSMNBench. Task abbreviations: \textbf{Act.} (Fine-grained Action), \textbf{H-H} (Human-Human Contact), \textbf{H-O} (Human-Object Contact), \textbf{Self} (Relative Self-Pose), \textbf{Joint} (Anatomical Joint), \textbf{Count} (Distinct Human Counting), \textbf{Attr.} (Counting with Attribute), \textbf{Cross} (Relative Cross-Person Pose), \textbf{Dist.} (Relative Distance), \textbf{Ori.} (Global Orientation), \textbf{Depth} (Global Depth Ordering). ``--'' indicates the setting is not applicable. $\delta_{dis} \downarrow$ represents the overall Distraction Decay (lower is better). Best results are in \textbf{bold}.}
\label{tab:main_results_part1}
\resizebox{\textwidth}{!}{
\begin{tabular}{@{}ll ccccc|c cccccc|c | c@{}}
\toprule
\multirow{2}{*}{\textbf{Model}} & \multirow{2}{*}{\textbf{Setting}} & \multicolumn{6}{c}{\textbf{SVS Tasks}} & \multicolumn{7}{c}{\textbf{MVN Tasks}} & \multirow{2}{*}{$\delta_{dis} \downarrow$} \\
\cmidrule(lr){3-8} \cmidrule(lr){9-15}
& & Act. & H-H & H-O & Self & Joint & \textbf{Avg} & Count & Attr. & Cross & Dist. & Ori. & Depth & \textbf{Avg} & \\
\midrule

\multicolumn{16}{@{}l}{\textbf{Baselines}} \\
\midrule
Random Guess      & -- &  24.0  &  24.7  &  24.0  &  26.3  &  26.0  &  25.0  &  25.3  &  24.3  &  25.3  &  23.7  &  24.3  &  28.0  &  25.2  &  --  \\
Blind             & -- &  27.0  &  22.3  &  26.3  &  29.3  &  29.7  &  26.9  &  21.3  &  25.0  &  22.7  &  29.7  &  21.0  &  28.0  &  24.6  &  --  \\
Human             & -1 (MVN only) &  --  &  --  &  --  &  --  &  --  &  --  &  52.3  &  58.7  &  62.0  &  65.7  &  64.0  &  68.3  &  61.8  &  --  \\
Human             & +0 (GT views) &  92.3  &  93.7  &  94.0  &  93.3  &  95.7  &  93.8  &  90.3  &  93.3  &  91.7  &  94.3  &  93.0  &  95.3  &  93.0  &  --  \\
Human             & +2 views &  93.0  &  93.3  &  94.3  &  92.0  &  94.0  &  93.3  &  87.0  &  91.7  &  92.3  &  93.7  &  92.3  &  91.8  &  92.5  &  --  \\
\midrule

\multicolumn{16}{@{}l}{\textbf{Proprietary Models}} \\
\midrule
\multirow{5}{*}{\begin{tabular}[c]{@{}l@{}}Gemini-2.5- \\ Flash\end{tabular}} 
& -1 (MVN only) & -- & -- & -- & -- & -- & -- & 21.3 & 34.7 & 45.0 & 48.7 & 29.0 & 36.7 & 35.9 & \multirow{5}{*}{2.7} \\
& +0 (GT views) & 36.3 & 42.3 & 44.3 & 46.7 & 44.0 & 42.7 & 39.7 & 39.7 & 42.7 & 41.3 & \textbf{39.0} & 34.3 & 39.4 & \\
& +1 view       & 38.7 & 32.0 & 41.3 & 44.0 & 42.7 & 39.7 & 33.3 & 36.0 & 46.7 & 45.0 & 24.7 & 36.0 & 36.9 & \\
& +2 views      & 33.7 & 31.7 & 41.7 & 42.3 & 44.7 & 38.8 & 35.0 & 39.3 & 46.0 & 45.3 & 28.7 & 41.3 & 39.3 & \\
& +3 (SVS only) & 33.3 & 27.3 & 40.0 & 44.7 & 41.3 & 37.3 & -- & -- & -- & -- & -- & -- & -- & \\
\midrule

\multirow{5}{*}{\begin{tabular}[c]{@{}l@{}}Gemini-2.5- \\ Pro\end{tabular}} 
& -1 (MVN only) & -- & -- & -- & -- & -- & -- & 27.3 & 38.7 & 49.7 & 53.7 & 34.7 & \textbf{47.0} & 41.9 & \multirow{5}{*}{2.3} \\
& +0 (GT views) & 38.7 & \textbf{46.7} & 47.7 & 46.7 & 43.0 & 44.6 & 43.7 & \textbf{46.7} & 45.0 & 48.7 & 33.0 & 46.7 & 44.0 & \\
& +1 view       & 34.0 & 38.7 & 39.7 & 47.7 & 42.0 & 40.4 & 45.7 & 41.7 & \textbf{51.7} & \textbf{56.0} & 31.3 & 46.7 & \textbf{45.5} & \\
& +2 views      & 35.3 & 37.7 & 40.0 & 47.3 & 41.3 & 40.3 & 36.3 & 42.3 & 49.0 & 47.7 & 27.7 & 46.7 & 41.6 & \\
& +3 (SVS only) & 36.0 & 38.3 & 40.7 & 48.7 & 40.3 & 40.8 & -- & -- & -- & -- & -- & -- & -- & \\
\midrule

\multirow{5}{*}{GPT-5.2} 
& -1 (MVN only) & -- & -- & -- & -- & -- & -- & 25.3 & 33.0 & 48.7 & 52.7 & 30.3 & 38.7 & 38.1 & \multirow{5}{*}{3.7} \\
& +0 (GT views) & 40.3 & 44.3 & 48.0 & \textbf{55.3} & \textbf{49.7} & \textbf{47.5} & 44.0 & 41.7 & 51.0 & 48.7 & 32.3 & 37.0 & 42.4 & \\
& +1 view       & \textbf{42.3} & 40.0 & 48.3 & 51.7 & 48.3 & 46.1 & 39.7 & 39.3 & 44.7 & 48.0 & 25.3 & 33.7 & 38.4 & \\
& +2 views      & 38.7 & 42.7 & 43.7 & 53.3 & 46.3 & 44.9 & 34.7 & 36.0 & 45.7 & 48.7 & 23.3 & 37.0 & 37.6 & \\
& +3 (SVS only) & 40.3 & 40.3 & 45.0 & 42.3 & 44.3 & 42.4 & -- & -- & -- & -- & -- & -- & -- & \\
\midrule

\multicolumn{16}{@{}l}{\textbf{Open-source Models}} \\
\midrule
\multirow{5}{*}{Qwen3-32B} 
& -1 (MVN only) & -- & -- & -- & -- & -- & -- & 28.3 & 34.3 & 37.3 & 47.0 & 27.7 & 39.0 & 35.6 & \multirow{5}{*}{1.7} \\
& +0 (GT views) & 38.7 & 38.0 & 46.3 & 45.3 & 35.0 & 40.7 & 42.0 & 41.7 & 41.0 & 48.7 & 26.7 & 40.7 & 40.1 & \\
& +1 view       & 36.7 & 36.7 & 43.7 & 42.0 & 33.7 & 38.6 & 42.7 & 43.3 & 41.0 & 49.0 & 27.0 & 38.7 & 40.3 & \\
& +2 views      & 35.0 & 36.0 & 42.3 & 40.0 & 33.0 & 37.3 & 41.0 & 40.3 & 41.0 & 49.0 & 21.0 & 42.0 & 39.1 & \\
& +3 (SVS only) & 34.7 & 36.3 & 43.3 & 40.3 & 31.7 & 37.3 & -- & -- & -- & -- & -- & -- & -- & \\
\midrule

\multirow{5}{*}{Qwen2.5-7B} 
& -1 (MVN only) & -- & -- & -- & -- & -- & -- & 21.3 & 32.0 & 32.7 & 42.7 & 22.3 & 35.0 & 31.0 & \multirow{5}{*}{0.9} \\
& +0 (GT views) & 33.3 & 28.3 & 36.0 & 31.3 & 27.7 & 31.3 & 32.0 & 35.0 & 35.0 & 43.7 & 24.3 & 42.0 & 35.3 & \\
& +1 view       & 33.7 & 29.7 & 37.7 & 32.0 & 28.3 & 32.3 & 36.3 & 33.3 & 34.0 & 39.0 & 21.7 & 42.7 & 34.5 & \\
& +2 views      & 28.3 & 29.3 & 37.3 & 30.3 & 30.0 & 31.0 & 34.0 & 29.0 & 35.0 & 34.3 & 22.3 & 40.7 & 32.6 & \\
& +3 (SVS only) & 30.0 & 28.3 & 34.7 & 31.3 & 29.0 & 30.7 & -- & -- & -- & -- & -- & -- & -- & \\
\midrule

\multirow{5}{*}{Qwen2.5-72B} 
& -1 (MVN only) & -- & -- & -- & -- & -- & -- & 26.3 & 32.0 & 40.0 & 51.0 & 28.3 & 35.0 & 35.4 & \multirow{5}{*}{2.5} \\
& +0 (GT views) & 39.0 & 41.7 & \textbf{49.3} & 44.3 & 39.0 & 42.7 & 41.3 & 36.0 & 39.0 & 48.0 & 28.3 & 37.0 & 38.3 & \\
& +1 view       & 39.7 & 42.7 & 45.7 & 39.7 & 39.3 & 41.4 & 39.0 & 36.3 & 36.7 & 44.0 & 23.0 & 36.0 & 35.8 & \\
& +2 views      & 36.7 & 38.3 & 43.3 & 40.3 & 35.3 & 38.8 & 41.3 & 39.3 & 34.0 & 37.7 & 29.0 & 35.7 & 36.2 & \\
& +3 (SVS only) & 39.0 & 39.7 & 43.3 & 38.0 & 39.0 & 39.8 & -- & -- & -- & -- & -- & -- & -- & \\
\midrule

\multirow{5}{*}{\begin{tabular}[c]{@{}l@{}}InternVL3.5- \\ 38B\end{tabular}} 
& -1 (MVN only) & -- & -- & -- & -- & -- & -- & 33.7 & 31.7 & 32.0 & 50.0 & 25.7 & 31.7 & 34.1 & \multirow{5}{*}{\textbf{0.5}} \\
& +0 (GT views) & 32.7 & 24.7 & 40.3 & 40.0 & 35.0 & 34.5 & 28.7 & 35.3 & 32.7 & 45.7 & 25.3 & 31.0 & 33.1 & \\
& +1 view       & 34.3 & 24.7 & 41.7 & 39.7 & 34.3 & 34.9 & 32.3 & 34.7 & 30.7 & 45.7 & 25.3 & 29.0 & 33.0 & \\
& +2 views      & 32.3 & 27.7 & 41.3 & 38.0 & 34.7 & 34.8 & 30.0 & 29.0 & 30.7 & 46.0 & 22.7 & 29.0 & 31.2 & \\
& +3 (SVS only) & 33.3 & 25.7 & 41.7 & 35.7 & 33.0 & 33.9 & -- & -- & -- & -- & -- & -- & -- & \\
\midrule

\multirow{5}{*}{\begin{tabular}[c]{@{}l@{}}InternVL3.5- \\ 78B\end{tabular}} 
& -1 (MVN only) & -- & -- & -- & -- & -- & -- & 28.0 & 30.3 & 37.0 & 51.0 & 31.0 & 42.0 & 36.6 & \multirow{5}{*}{1.7} \\
& +0 (GT views) & 34.0 & 30.3 & 38.7 & 40.3 & 37.7 & 36.2 & \textbf{46.0} & 40.7 & 39.0 & 44.0 & 27.3 & 38.7 & 39.3 & \\
& +1 view       & 33.7 & 28.7 & 38.0 & 38.7 & 35.3 & 34.9 & 42.3 & 41.0 & 39.0 & 45.0 & 27.3 & 36.0 & 38.4 & \\
& +2 views      & 32.0 & 28.0 & 37.7 & 39.7 & 34.3 & 34.3 & 40.3 & 40.3 & 39.7 & 41.7 & 26.0 & 33.7 & 36.9 & \\
& +3 (SVS only) & 31.3 & 27.7 & 37.3 & 38.0 & 35.7 & 34.0 & -- & -- & -- & -- & -- & -- & -- & \\
\bottomrule

\end{tabular}
}
\end{table}

\subsection{Main Results}
\label{sec:main_results}
\vspace{-0.5em}

Based on extensive evaluations across the SVS and MVN subsets in Table~\ref{tab:main_results_part1}, we synthesize our empirical findings and architectural analyses into \textbf{five} key insights detailing the capabilities and limitations of MLLMs in complex cross-view human and human-object understanding.

\vspace{0.2em}
\noindent\textcolor{teal}{$\blacktriangleright$~\textbf{4.2.1 Current MLLMs struggle fundamentally, despite proprietary leadership and specific open-source strengths.}}
Despite recent advancements, a severe performance gap persists between MLLMs and human perception. Human evaluators consistently exceed 87\% accuracy in both the \textit{+0} and \textit{+2} settings. Even in the \textit{-1} setting, humans maintain robust performance by systematically eliminating incorrect options and inferring the most probable answer from partial evidence. In contrast, top proprietary models (\eg, GPT-5.2) peak at only 47.5\% (SVS) and 42.4\% (MVN), barely about 15\% above Random Guess and Blind baselines. While proprietary models generally dominate spatial reasoning tasks due to massive parameter scales, high-capacity open-source models demonstrate strong task-specific competitiveness. For instance, Qwen2.5-72B achieves the highest accuracy in SVS Human-Object Contact (49.3\%), narrowly outperforming GPT-5.2 (48.0\%). These findings underscore that achieving robust, human-level cross-view understanding in fine-grained, occlusion-heavy scenes remains a critical open challenge.

\vspace{0.2em}
\noindent\textcolor{teal}{$\blacktriangleright$~\textbf{4.2.2 Additional visual inputs consistently trigger a universal distraction degradation phenomenon.}}
The evaluation explicitly exposes the brittleness of MLLMs' selective visual attention across both SVS and MVN tasks. Theoretically, providing additional views (\textit{+1, +2, +3}) alongside the necessary ground-truth views (\textit{+0}) should not harm decision-making. However, the Distraction Decay ($\delta_{dis}$) metric reveals a consistent performance drop across nearly all models as more views are introduced. This empirically validates that modern MLLMs process multiple images by loosely averaging semantic features rather than dynamically isolating the most informative viewpoints, leading to context saturation and confusion when exposed to uninformative angles.


\vspace{0.2em}
\noindent\textcolor{teal}{$\blacktriangleright$~\textbf{4.2.3 Higher model capacity paradoxically increases vulnerability to visual distraction.}}
An analysis of the $\delta_{dis}$ metric reveals an intriguing paradox: models with higher overall capabilities often exhibit greater vulnerability to visual noise. High-capacity proprietary models like GPT-5.2 and Gemini-2.5-Flash exhibit higher decay rates (3.7 and 2.7, respectively), whereas smaller or lower-performing open-source models like InternVL3.5-38B show a remarkable resilience to distraction ($\delta_{dis}$ of 0.5), albeit at a much lower baseline accuracy ($\sim$33\%). This suggests that more parameter-rich models aggressively attempt to extract cross-attention patterns from all provided visual tokens, making their reasoning pathways more easily hijacked by irrelevant visual distractors.

\vspace{0.2em}
\noindent\textcolor{teal}{$\blacktriangleright$~\textbf{4.2.4 The prevailing bag-of-frames input paradigm fails to facilitate true cross-view understanding.}}
The benchmark results challenge the prevailing ``bag of frames'' paradigm in multi-image MLLM evaluation. The systematic perturbation of view availability proves that simply feeding an architecture more viewpoints does not equate to a deeper understanding of the scene. Because current visual encoders flatten independent 2D frames into a 1D token sequence without strict epistemic or geometric grounding, models lack the inductive biases required to map shared entities across different coordinate spaces. Consequently, the empirical evidence indicates that merely increasing the quantity of 2D multi-image inputs does not intrinsically translate into robust cross-view spatial comprehension.

\begin{table}[t]
\centering
\caption{Ablation on different input resolution using Gemini-2.5-Flash. Task abbreviations are the same as Table~\ref{tab:main_results_part1}. Best results are in \textbf{bold}.}
\vspace{-0.5em}
\label{tab:ablation}
\resizebox{\textwidth}{!}{
\begin{tabular}{@{}ll ccccc|c cccccc|c | c@{}}
\toprule
\multirow{2}{*}{\textbf{Input Setting}} & \multirow{2}{*}{\textbf{View Setting}} & \multicolumn{6}{c}{\textbf{SVS Tasks}} & \multicolumn{7}{c}{\textbf{MVN Tasks}} & \multirow{2}{*}{$\delta_{dis} \downarrow$} \\
\cmidrule(lr){3-8} \cmidrule(lr){9-15}
& & Act. & H-H & H-O & Self & Joint & \textbf{Avg} & Count & Attr. & Cross & Dist. & Ori. & Depth & \textbf{Avg} & \\
\midrule

\multirow{5}{*}{1920$\times$1080} 
& -1 (MVN only) & -- & -- & -- & -- & -- & -- & 21.3 & 34.7 & 45.0 & 48.7 & 29.0 & 36.7 & 35.9 & \multirow{5}{*}{2.7} \\
& +0 (GT views) & 36.3 & \textbf{42.3} & 44.3 & \textbf{46.7} & \textbf{44.0} & 42.7 & \textbf{39.7} & 39.7 & 42.7 & 41.3 & \textbf{39.0} & 34.3 & \textbf{39.4} & \\
& +1 view       & 38.7 & 32.0 & 41.3 & 44.0 & 42.7 & 39.7 & 33.3 & 36.0 & 46.7 & 45.0 & 24.7 & 36.0 & 36.9 & \\
& +2 views      & 33.7 & 31.7 & 41.7 & 42.3 & 44.7 & 38.8 & 35.0 & 39.3 & 46.0 & 45.3 & 28.7 & 41.3 & 39.3 & \\
& +3 (SVS only) & 33.3 & 27.3 & 40.0 & 44.7 & 41.3 & 37.3 & -- & -- & -- & -- & -- & -- & -- & \\
\midrule

\multirow{5}{*}{1280$\times$720} 
& -1 (MVN only) & -- & -- & -- & -- & -- & -- & 21.7 & 36.3 & 42.7 & 49.0 & 33.3 & \textbf{47.0} & 38.3 & \multirow{5}{*}{\textbf{2.2}} \\
& +0 (GT views) & 37.7 & 39.7 & 44.7 & 44.3 & 42.7 & 41.8 & 37.0 & 39.7 & 45.0 & 42.7 & 28.7 & 34.3 & 37.9 & \\
& +1 view       & 37.0 & 31.7 & 41.3 & 41.7 & 39.0 & 38.1 & 35.7 & 38.7 & 46.7 & 44.0 & 31.3 & 35.0 & 38.6 & \\
& +2 views      & 34.0 & 31.3 & 41.7 & 39.7 & 37.7 & 36.9 & 30.0 & 39.0 & 45.7 & 39.7 & 27.0 & 40.7 & 37.0 & \\
& +3 (SVS only) & 35.7 & 35.3 & 38.3 & 37.7 & 40.3 & 37.5 & -- & -- & -- & -- & -- & -- & -- & \\
\midrule

\multirow{5}{*}{640$\times$480} 
& -1 (MVN only) & -- & -- & -- & -- & -- & -- & 21.3 & 30.7 & 41.0 & 48.0 & 34.7 & \textbf{47.0} & 37.1 & \multirow{5}{*}{2.3} \\
& +0 (GT views) & \textbf{41.7} & 38.0 & 42.7 & 42.7 & 43.7 & 41.8 & 32.7 & 35.0 & 45.0 & 45.7 & 28.7 & 38.0 & 37.5 & \\
& +1 view       & 35.3 & 33.7 & 40.0 & 44.0 & 36.0 & 37.8 & 36.7 & 35.3 & 43.0 & 44.0 & 30.0 & 35.0 & 37.3 & \\
& +2 views      & 37.3 & 32.7 & 38.0 & 41.3 & 38.0 & 37.5 & 30.3 & 36.0 & 47.0 & 44.0 & 28.7 & 33.0 & 36.5 & \\
& +3 (SVS only) & 38.0 & 35.3 & 39.3 & 40.0 & 39.0 & 38.3 & -- & -- & -- & -- & -- & -- & -- & \\
\midrule

\multirow{5}{*}{fixed} 
& -1 (MVN only) & -- & -- & -- & -- & -- & -- & 22.7 & 36.7 & 41.7 & \textbf{51.7} & 34.7 & 45.7 & 38.9 & \multirow{5}{*}{4.0} \\
& +0 (GT views) & 39.0 & \textbf{44.3} & \textbf{45.7} & 45.3 & 39.3 & \textbf{42.7} & 36.0 & \textbf{40.7} & 45.0 & 44.7 & 30.0 & 40.0 & \textbf{39.4} & \\
& +1 view       & 33.0 & 32.0 & 39.7 & 36.7 & 37.3 & 35.7 & 34.0 & 36.7 & \textbf{47.0} & 45.0 & 26.0 & 38.7 & 37.9 & \\
& +2 views      & 38.7 & 31.3 & 36.7 & 42.7 & 37.7 & 37.4 & 32.3 & 38.0 & 46.0 & 37.7 & 30.3 & 34.3 & 36.4 & \\
& +3 (SVS only) & 37.0 & 32.0 & 44.7 & 38.7 & 37.0 & 37.9 & -- & -- & -- & -- & -- & -- & -- & \\
\bottomrule

\end{tabular}
}
\vspace{-1.2em}
\end{table}

\vspace{0.2em}
\noindent\textcolor{teal}{$\blacktriangleright$~\textbf{4.2.5 Multi-view fusion exposes a viewpoint conflict dilemma and an overreliance on monocular priors.}}
The systematic removal of a necessary view (the -1 setting) reveals a distinct dichotomy in MVN tasks. While omitting a view predictably degrades entity-counting performance due to missing information, spatial tasks exhibit a counterintuitive trend: models frequently match or surpass full-context (+0) performance in the view-depleted -1 setting. This phenomenon highlights a fundamental architectural flaw. Restricted to a single viewpoint, models leverage strong monocular priors from large-scale 2D pre-training to generate plausible spatial estimates. However, introducing supplementary views triggers representational conflict. Because physical entities exhibit drastically different 2D projections across cameras, standard 1D Transformer attention mechanisms struggle to reconcile them through geometric triangulation. Rather than synthesizing complementary perspectives, models process them as mutually interfering signals. Consequently, providing the exact geometric data needed to resolve ambiguity actively undermines the model's single-image baseline, proving that current architectures intrinsically treat multi-view inputs as visual distractions rather than collaborative spatial cues.

\vspace{-1em}
\subsection{Impact of Input Resolution}
\label{sce:ablation}
\vspace{-0.5em}
To investigate the impact of visual input scale on multi-view reasoning, we ablate Gemini-2.5-Flash under two constraints: (1) \textbf{Different Resolutions:} Scaling images to 1920$\times$1080, 1280$\times$720, and 640$\times$480. (2) \textbf{Fixed Total Resolution:} Maintaining a constant $1920 \times 1080$ pixel count for the input window, dynamically resizing individual frames based on the view count.

Our findings (Table~\ref{tab:ablation}) confirm that high input resolution is essential for fine-grained perception and distraction robustness. The 1080P setting achieves peak baseline (\textit{+0}) accuracy (SVS: 42.7\%, MVN: 39.4\%). Degrading to 720P or 480P consistently impairs multi-view fusion and nearly doubles the distraction degradation seen at 1080P ($\delta_{dis} = 1.2$). This proves that compromised visual clarity directly exacerbates vulnerability to uninformative views, showing that high-fidelity inputs are required to maintain stable spatial reasoning.

Conversely, the ``fixed'' total resolution strategy, a common context-saving optimization, harms spatial reasoning. Although its baseline mirrors 1080P, accuracy collapses as additional views force progressive frame downscaling. This continuous loss of per-frame clarity yields the highest Distraction Decay ($\delta_{dis} = 4.0$), proving that dynamic image compression artificially degrades multi-view performance and that maintaining per-frame fidelity is critical when scaling inputs.

\begin{figure}[tb]
  \centering
  \includegraphics[height=5.5cm]{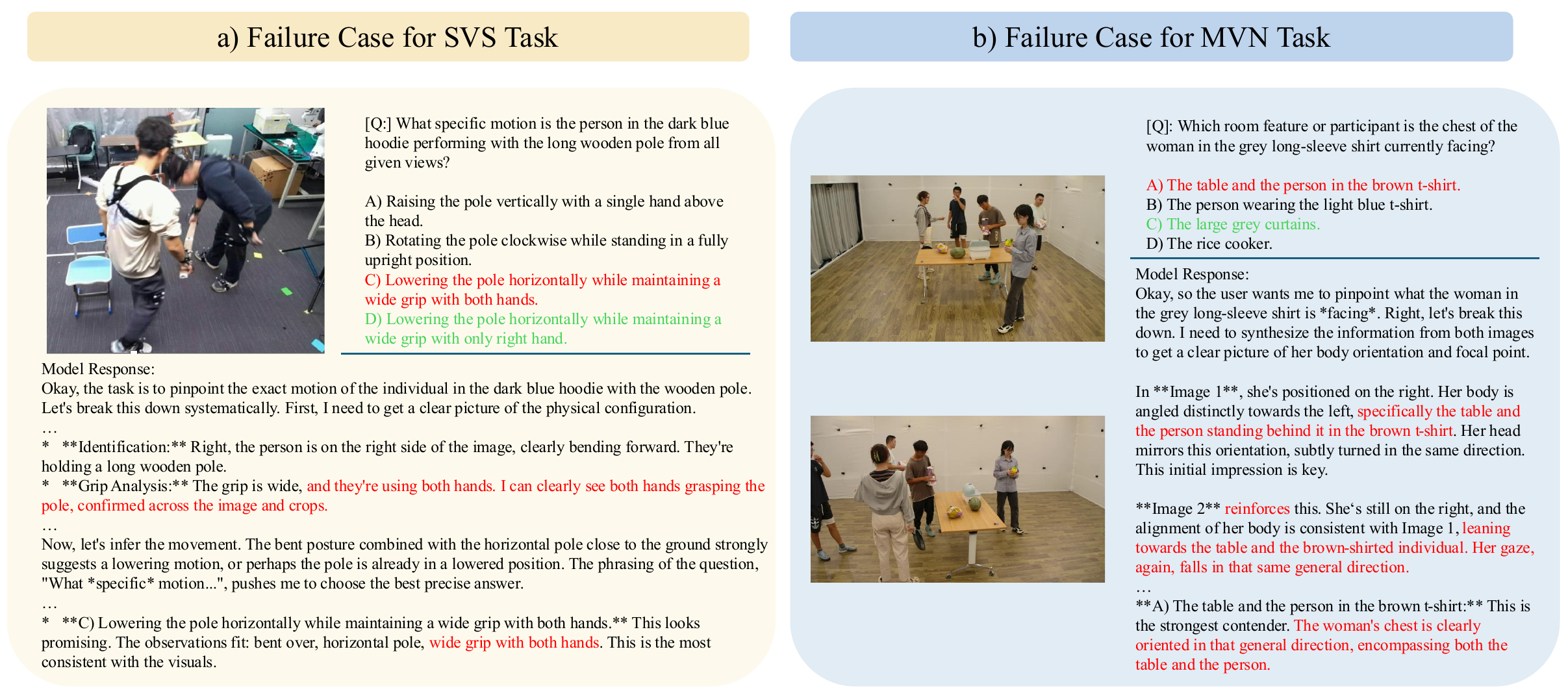}
  \vspace{-0.7em}
  \caption{Qualitative examples of typical model failure cases. (a) In the SVS task, the model hallucinates fine-grained interaction details, incorrectly concluding that the subject is maintaining a grip with both hands. (b) In the MVN task, the model fails to synthesize cross-view geometric evidence, incorrectly determining the subject's global orientation by over-relying on the deceptive perspective from a single view.}
  \label{fig:failure_case}
    \vspace{-1.0em}

\end{figure}

\vspace{-0.5em}
\section{Failure Case Analysis}
\label{sec:error_analysis}
\vspace{-0.5em}

We perform a large-scale, systematic manual error analysis on Gemini-2.5-Pro, GPT-5.2, and Qwen2.5-72B. Using stratified random sampling, we select 60 failed instances per question type across the 11 tasks, resulting in 660 manually audited questions. We group the dominant failure modes into two axes: \textbf{Image-level errors} (failures in parsing individual 2D frames) and \textbf{View-level errors} (failures in multi-frame fusion). The quantitative results are reported in Table~\ref{tab:failure_cases}, and qualitative examples are illustrated in Figure~\ref{fig:failure_case}.

\begin{table}[htbp]
  \centering
  \caption{Distribution of manually audited failure cases categorized by image-level and view-level errors.}
  \label{tab:failure_cases}
  \resizebox{0.85\textwidth}{!}{
  \begin{tabular}{@{}llc@{}}
    \toprule
    \textbf{Error Dimension} & \textbf{Detailed Sub-Category} & \textbf{Ratio (\%)} \\
    \midrule
    \multirow{5}{*}{\textbf{Image-Level}} 
    & Spatial Analysis Failure (Height, Direction, Depth) & 33.1 \\
    & Contact Area Grounding (Human-Object / Human-Human) & 21.1 \\
    & Fine-Grained Human Joint Recognition (Action \& Status) & 19.3 \\
    & Cross-Image Entity Linking \& Disambiguation & 14.7 \\
    & Partial Visibility \& Appearance Recognition & 12.8 \\
    \midrule
    \multirow{2}{*}{\textbf{View-Level}} 
    & Conflict Information Awareness \& Fusion Failure & 67.2 \\
    & Over-Reliance on Preferred View (Semantic Override) & 32.8 \\
    \bottomrule
  \end{tabular}
  }
\end{table}

\noindent\textcolor{teal}{$\blacktriangleright$~\textbf{Image-Level Errors.}}
These fundamental 2D perception failures occur independently of cross-view integration. The primary bottleneck is \textbf{Spatial Analysis Failure (33.1\%)}, where models fail to infer 3D relationships and depth. \textbf{Contact Area Grounding (21.1\%)} errors cause hallucinated interactions due to poor boundary localization. \textbf{Fine-Grained Joint Recognition (19.3\%)} reveals brittle micro-level pose understanding (\eg, misjudging flexion). Finally, \textbf{Cross-Image Entity Linking (14.7\%)} and \textbf{Partial Visibility (12.8\%)} expose models' inability to track instances across coordinate spaces or recognize heavily truncated subjects.

\noindent\textcolor{teal}{$\blacktriangleright$~\textbf{View-Level Errors.}}
These errors expose brittle multi-image attention during collaborative evidence synthesis. The dominant flaw, \textbf{Conflict Info Awareness \& Fusion Failure (67.2\%)}, occurs when models detect conflicting cross-view cues but fail to integrate the fragmented details. Conversely, \textbf{Over-Reliance on Preferred View (32.8\%)} reveals severe positional or semantic bias: rather than synthesizing all perspectives, models anchor on a single ``preferred'' view and forcibly apply its observation globally, effectively ignoring secondary frames.

\vspace{-1em}
\section{Conclusion}
\label{sec:conclusion}
\vspace{-0.5em}
To summarize, we introduce SSMNBench, a diagnostic benchmark specifically designed to assess MLLMs on complex cross-view human-centric understanding. Our evaluation across 17 state-of-the-art models reveals a substantial gap between current architectures and human-level spatial reasoning. By categorizing tasks into Single-View Sufficiency (SVS) and Multi-View Necessity (MVN) under perturbed view availability, we distinguish distraction robustness from genuine multi-view fusion, highlighting that models suffer severe performance decay from redundant views and fail to integrate fragmented geometric evidence. We hope SSMNBench will serve as a valuable resource for the community and advance the progress toward true cross-view synthesis in future multimodal AI systems.

\vspace{1em}
\noindent\textbf{Acknowledgements.} This research is funded in part by ARC-Discovery grant (DP220100800 to XY), ARC-DECRA grant (DE230100477 to XY), the Advance Queensland Industry Research Projects (AQIRP), and Follow ME PTY LTD. We thank all anonymous reviewers and ACs for their constructive suggestions.

%
%
\bibliographystyle{splncs04}
\bibliography{main}

@String(CVPR  = {IEEE Conf. Comput. Vis. Pattern Recog.})

@String(ICCV  = {Int. Conf. Comput. Vis.})

@String(NeurIPS = {Adv. Neural Inform. Process. Syst.})

@String(CVPR  = {CVPR})

@String(ICCV  = {ICCV})

@String(NeurIPS = {NeurIPS})

@inproceedings{yang2025thinking,
  title={Thinking in space: How multimodal large language models see, remember, and recall spaces},
  author={Yang, Jihan and Yang, Shusheng and Gupta, Anjali W and Han, Rilyn and Fei-Fei, Li and Xie, Saining},
  booktitle={Proceedings of the Computer Vision and Pattern Recognition Conference},
  pages={10632--10643},
  year={2025}
}

@article{yeh2025seeing,
  title={Seeing from another perspective: Evaluating multi-view understanding in mllms},
  author={Yeh, Chun-Hsiao and Wang, Chenyu and Tong, Shengbang and Cheng, Ta-Ying and Wang, Ruoyu and Chu, Tianzhe and Zhai, Yuexiang and Chen, Yubei and Gao, Shenghua and Ma, Yi},
  journal={arXiv preprint arXiv:2504.15280},
  year={2025}
}

@misc{gholami2025spatialreasoningvisionlanguagemodels,
      title={Spatial Reasoning with Vision-Language Models in Ego-Centric Multi-View Scenes}, 
      author={Mohsen Gholami and Ahmad Rezaei and Zhou Weimin and Sitong Mao and Shunbo Zhou and Yong Zhang and Mohammad Akbari},
      year={2025},
      eprint={2509.06266},
      archivePrefix={arXiv},
      primaryClass={cs.CV},
      url={https://arxiv.org/abs/2509.06266}, 
}

@article{wang2024muirbench,
  title={Muirbench: A comprehensive benchmark for robust multi-image understanding},
  author={Wang, Fei and Fu, Xingyu and Huang, James Y and Li, Zekun and Liu, Qin and Liu, Xiaogeng and Ma, Mingyu Derek and Xu, Nan and Zhou, Wenxuan and Zhang, Kai and others},
  journal={arXiv preprint arXiv:2406.09411},
  year={2024}
}

@inproceedings{fu2024blink,
  title={Blink: Multimodal large language models can see but not perceive},
  author={Fu, Xingyu and Hu, Yushi and Li, Bangzheng and Feng, Yu and Wang, Haoyu and Lin, Xudong and Roth, Dan and Smith, Noah A and Ma, Wei-Chiu and Krishna, Ranjay},
  booktitle={European Conference on Computer Vision},
  pages={148--166},
  year={2024},
  organization={Springer}
}

@article{yang2025mmsi,
  title={Mmsi-bench: A benchmark for multi-image spatial intelligence},
  author={Yang, Sihan and Xu, Runsen and Xie, Yiman and Yang, Sizhe and Li, Mo and Lin, Jingli and Zhu, Chenming and Chen, Xiaochen and Duan, Haodong and Yue, Xiangyu and others},
  journal={arXiv preprint arXiv:2505.23764},
  year={2025}
}

@inproceedings{li2024mvbench,
  title={Mvbench: A comprehensive multi-modal video understanding benchmark},
  author={Li, Kunchang and Wang, Yali and He, Yinan and Li, Yizhuo and Wang, Yi and Liu, Yi and Wang, Zun and Xu, Jilan and Chen, Guo and Luo, Ping and others},
  booktitle={Proceedings of the IEEE/CVF Conference on Computer Vision and Pattern Recognition},
  pages={22195--22206},
  year={2024}
}

@inproceedings{grauman2024ego,
  title={Ego-exo4d: Understanding skilled human activity from first-and third-person perspectives},
  author={Grauman, Kristen and Westbury, Andrew and Torresani, Lorenzo and Kitani, Kris and Malik, Jitendra and Afouras, Triantafyllos and Ashutosh, Kumar and Baiyya, Vijay and Bansal, Siddhant and Boote, Bikram and others},
  booktitle={Proceedings of the IEEE/CVF Conference on Computer Vision and Pattern Recognition},
  pages={19383--19400},
  year={2024}
}

@InProceedings{Zhu_2023_ICCV,
    author    = {Zhu, Yue and Samet, Nermin and Picard, David},
    title     = {H3WB: Human3.6M 3D WholeBody Dataset and Benchmark},
    booktitle = {Proceedings of the IEEE/CVF International Conference on Computer Vision (ICCV)},
    month     = {October},
    year      = {2023},
    pages     = {20166-20177}
}

@inproceedings{xu2025m3gym,
  title={M3GYM: A Large-Scale Multimodal Multi-view Multi-person Pose Dataset for Fitness Activity Understanding in Real-world Settings},
  author={Xu, Qingzheng and Cao, Ru and Shen, Xin and Du, Heming and Wang, Sen and Yu, Xin},
  booktitle={Proceedings of the Computer Vision and Pattern Recognition Conference},
  pages={12289--12300},
  year={2025}
}

@inproceedings{ozsoy20224d,
  title={4d-or: Semantic scene graphs for or domain modeling},
  author={{\"O}zsoy, Ege and {\"O}rnek, Evin P{\i}nar and Eck, Ulrich and Czempiel, Tobias and Tombari, Federico and Navab, Nassir},
  booktitle={International conference on medical image computing and computer-assisted intervention},
  pages={475--485},
  year={2022},
  organization={Springer}
}

@inproceedings{ozsoy2024mmor,
  title={MM-OR: A Large Multimodal Operating Room Dataset for Semantic Understanding of High Intensity Surgical Environments},
  author={Ege Özsoy and Pellegrini, Chantal and Czempiel, Tobias and Tristram, Felix and Yuan, Kun and Bani-Harouni, David and Eck, Ulrich and Busam, Benjamin and Keicher, Matthias and Navab, Nassir},
  booktitle={CVPR},
  note={Accepted},
  year={2025}
}

@inproceedings{gan2021mvmhat,
  title={Self-supervised multi-view multi-human association and tracking},
  author={Gan, Yiyang and Han, Ruize and Yin, Liqiang and Feng, Wei and Wang, Song},
  booktitle={Proceedings of the 29th ACM international conference on multimedia},
  pages={282--290},
  year={2021}
}

@article{khirodkar2024harmony4d,
  title={Harmony4d: A video dataset for in-the-wild close human interactions},
  author={Khirodkar, Rawal and Song, Jyun-Ting and Cao, Jinkun and Luo, Zhengyi and Kitani, Kris},
  journal={Advances in Neural Information Processing Systems},
  volume={37},
  pages={107270--107285},
  year={2024}
}

@inproceedings{zhang2024hoi,
  title={Hoi-m\^{} 3: Capture multiple humans and objects interaction within contextual environment},
  author={Zhang, Juze and Zhang, Jingyan and Song, Zining and Shi, Zhanhe and Zhao, Chengfeng and Shi, Ye and Yu, Jingyi and Xu, Lan and Wang, Jingya},
  booktitle={Proceedings of the IEEE/CVF Conference on Computer Vision and Pattern Recognition},
  pages={516--526},
  year={2024}
}

@inproceedings{khirodkar2023ego,
  title={Ego-humans: An ego-centric 3d multi-human benchmark},
  author={Khirodkar, Rawal and Bansal, Aayush and Ma, Lingni and Newcombe, Richard and Vo, Minh and Kitani, Kris},
  booktitle={Proceedings of the IEEE/CVF International Conference on Computer Vision},
  pages={19807--19819},
  year={2023}
}

@inproceedings{liu2025core4d,
  title={Core4d: A 4d human-object-human interaction dataset for collaborative object rearrangement},
  author={Liu, Yun and Zhang, Chengwen and Xing, Ruofan and Tang, Bingda and Yang, Bowen and Yi, Li},
  booktitle={Proceedings of the IEEE/CVF Conference on Computer Vision and Pattern Recognition},
  pages={1769--1782},
  year={2025}
}

@inproceedings{wu2023multimodal,
  title={Multimodal large language models: A survey},
  author={Wu, Jiayang and Gan, Wensheng and Chen, Zefeng and Wan, Shicheng and Yu, Philip S},
  booktitle={2023 IEEE International Conference on Big Data (BigData)},
  pages={2247--2256},
  year={2023},
  organization={IEEE}
}

@article{yin2024survey,
  title={A survey on multimodal large language models},
  author={Yin, Shukang and Fu, Chaoyou and Zhao, Sirui and Li, Ke and Sun, Xing and Xu, Tong and Chen, Enhong},
  journal={National Science Review},
  volume={11},
  number={12},
  pages={nwae403},
  year={2024},
  publisher={Oxford University Press}
}

@article{yang2025qwen3,
  title={Qwen3 technical report},
  author={Yang, An and Li, Anfeng and Yang, Baosong and Zhang, Beichen and Hui, Binyuan and Zheng, Bo and Yu, Bowen and Gao, Chang and Huang, Chengen and Lv, Chenxu and others},
  journal={arXiv preprint arXiv:2505.09388},
  year={2025}
}

@inproceedings{chen2024internvl,
  title={Internvl: Scaling up vision foundation models and aligning for generic visual-linguistic tasks},
  author={Chen, Zhe and Wu, Jiannan and Wang, Wenhai and Su, Weijie and Chen, Guo and Xing, Sen and Zhong, Muyan and Zhang, Qinglong and Zhu, Xizhou and Lu, Lewei and others},
  booktitle={Proceedings of the IEEE/CVF conference on computer vision and pattern recognition},
  pages={24185--24198},
  year={2024}
}

@article{bai2023qwen,
  title={Qwen technical report},
  author={Bai, Jinze and Bai, Shuai and Chu, Yunfei and Cui, Zeyu and Dang, Kai and Deng, Xiaodong and Fan, Yang and Ge, Wenbin and Han, Yu and Huang, Fei and others},
  journal={arXiv preprint arXiv:2309.16609},
  year={2023}
}

@misc{deepseekai2024deepseekv3technicalreport,
      title={DeepSeek-V3 Technical Report}, 
      author={DeepSeek-AI},
      year={2024},
      eprint={2412.19437},
      archivePrefix={arXiv},
      primaryClass={cs.CL},
      url={https://arxiv.org/abs/2412.19437}, 
}

@misc{liu2023llava,
      title={Visual Instruction Tuning}, 
      author={Liu, Haotian and Li, Chunyuan and Wu, Qingyang and Lee, Yong Jae},
      publisher={NeurIPS},
      year={2023},
}

@article{wang2025internvl3_5,
  title={InternVL3.5: Advancing Open-Source Multimodal Models in Versatility, Reasoning, and Efficiency},
  author={Wang, Weiyun and Gao, Zhangwei and Gu, Lixin and Pu, Hengjun and Cui, Long and Wei, Xingguang and Liu, Zhaoyang and Jing, Linglin and Ye, Shenglong and Shao, Jie and others},
  journal={arXiv preprint arXiv:2508.18265},
  year={2025}
}

@article{dubey2024llama,
  title={The llama 3 herd of models},
  author={Dubey, Abhimanyu and Jauhri, Abhinav and Pandey, Abhinav and Kadian, Abhishek and Al-Dahle, Ahmad and Letman, Aiesha and Mathur, Akhil and Schelten, Alan and Yang, Amy and Fan, Angela and others},
  journal={arXiv e-prints},
  pages={arXiv--2407},
  year={2024}
}

@article{comanici2025gemini,
  title={Gemini 2.5: Pushing the frontier with advanced reasoning, multimodality, long context, and next generation agentic capabilities},
  author={Comanici, Gheorghe and Bieber, Eric and Schaekermann, Mike and Pasupat, Ice and Sachdeva, Noveen and Dhillon, Inderjit and Blistein, Marcel and Ram, Ori and Zhang, Dan and Rosen, Evan and others},
  journal={arXiv preprint arXiv:2507.06261},
  year={2025}
}

@misc{qwen3.5,
    title  = {{Qwen3.5}: Towards Native Multimodal Agents},
    author = {{Qwen Team}},
    month  = {February},
    year   = {2026},
    url    = {https://qwen.ai/blog?id=qwen3.5}
}

@article{team2023gemini,
  title={Gemini: a family of highly capable multimodal models},
  author={Team, Gemini and Anil, Rohan and Borgeaud, Sebastian and Alayrac, Jean-Baptiste and Yu, Jiahui and Soricut, Radu and Schalkwyk, Johan and Dai, Andrew M and Hauth, Anja and Millican, Katie and others},
  journal={arXiv preprint arXiv:2312.11805},
  year={2023}
}

@misc{openai2025o3o4mini,
  author = {{OpenAI}},
  title = {Introducing GPT‑5.2},
  url = {https://openai.com/index/introducing-gpt-5-2/},
  year = {2025}
}

@misc{deepmind2025gemini25,
  author = {{Google DeepMind}},
  title = {Gemini 2.5: Our most intelligent ai model},
  url = {https://blog.google/technology/google-deepmind/gemini-model-thinking-updates-march-2025/},
  year = {2025}
}

@article{cho2025perceptionlm,
  title={Perceptionlm: Open-access data and models for detailed visual understanding},
  author={Cho, Jang Hyun and Madotto, Andrea and Mavroudi, Effrosyni and Afouras, Triantafyllos and Nagarajan, Tushar and Maaz, Muhammad and Song, Yale and Ma, Tengyu and Hu, Shuming and Jain, Suyog and others},
  journal={arXiv preprint arXiv:2504.13180},
  year={2025}
}

@article{shen2025vlm,
  title={Vlm-r1: A stable and generalizable r1-style large vision-language model},
  author={Shen, Haozhan and Liu, Peng and Li, Jingcheng and Fang, Chunxin and Ma, Yibo and Liao, Jiajia and Shen, Qiaoli and Zhang, Zilun and Zhao, Kangjia and Zhang, Qianqian and Xu, Ruochen and Zhao, Tiancheng },
  journal={arXiv preprint arXiv:2504.07615},
  year={2025}
}

@misc{chen2025r1v,
  author       = {Chen, Liang and Li, Lei and Zhao, Haozhe and Song, Yifan and Vinci},
  title        = {R1-V: Reinforcing Super Generalization Ability in Vision-Language Models with Less Than \$3},
  howpublished = {\url{https://github.com/Deep-Agent/R1-V}},
  note         = {Accessed: 2025-02-02},
  year         = {2025}
}

@misc{openr1,
    title = {Open R1: A fully open reproduction of DeepSeek-R1},
    url = {https://github.com/huggingface/open-r1},
    author = {{Hugging Face}},
    month = {January},
    year = {2025}
}

@misc{sun2025reinforcementfinetuningpowersreasoning,
      title={Reinforcement Fine-Tuning Powers Reasoning Capability of Multimodal Large Language Models}, 
      author={Haoyuan Sun and Jiaqi Wu and Bo Xia and Yifu Luo and Yifei Zhao and Kai Qin and Xufei Lv and Tiantian Zhang and Yongzhe Chang and Xueqian Wang},
      year={2025},
      eprint={2505.18536},
      archivePrefix={arXiv},
      primaryClass={cs.CL},
      url={https://arxiv.org/abs/2505.18536}, 
}

@misc{vteam2025glm45vglm41vthinkingversatilemultimodal,
      title={GLM-4.5V and GLM-4.1V-Thinking: Towards Versatile Multimodal Reasoning with Scalable Reinforcement Learning},
      author={V Team and Wenyi Hong and Wenmeng Yu and Xiaotao Gu and Guo Wang and Guobing Gan and Haomiao Tang and Jiale Cheng and Ji Qi and Junhui Ji and Lihang Pan and Shuaiqi Duan and Weihan Wang and Yan Wang and Yean Cheng and Zehai He and Zhe Su and Zhen Yang and Ziyang Pan and Aohan Zeng and Baoxu Wang and Bin Chen and Boyan Shi and Changyu Pang and Chenhui Zhang and Da Yin and Fan Yang and Guoqing Chen and Jiazheng Xu and Jiale Zhu and Jiali Chen and Jing Chen and Jinhao Chen and Jinghao Lin and Jinjiang Wang and Junjie Chen and Leqi Lei and Letian Gong and Leyi Pan and Mingdao Liu and Mingde Xu and Mingzhi Zhang and Qinkai Zheng and Sheng Yang and Shi Zhong and Shiyu Huang and Shuyuan Zhao and Siyan Xue and Shangqin Tu and Shengbiao Meng and Tianshu Zhang and Tianwei Luo and Tianxiang Hao and Tianyu Tong and Wenkai Li and Wei Jia and Xiao Liu and Xiaohan Zhang and Xin Lyu and Xinyue Fan and Xuancheng Huang and Yanling Wang and Yadong Xue and Yanfeng Wang and Yanzi Wang and Yifan An and Yifan Du and Yiming Shi and Yiheng Huang and Yilin Niu and Yuan Wang and Yuanchang Yue and Yuchen Li and Yutao Zhang and Yuting Wang and Yu Wang and Yuxuan Zhang and Zhao Xue and Zhenyu Hou and Zhengxiao Du and Zihan Wang and Peng Zhang and Debing Liu and Bin Xu and Juanzi Li and Minlie Huang and Yuxiao Dong and Jie Tang},
      year={2025},
      eprint={2507.01006},
      archivePrefix={arXiv},
      primaryClass={cs.CV},
      url={https://arxiv.org/abs/2507.01006},
}

@article{fu2023mme,
  title={Mme: A comprehensive evaluation benchmark for multimodal large language models},
  author={Fu, Chaoyou and Chen, Peixian and Shen, Yunhang and Qin, Yulei and Zhang, Mengdan and Lin, Xu and Yang, Jinrui and Zheng, Xiawu and Li, Ke and Sun, Xing and others},
  journal={arXiv preprint arXiv:2306.13394},
  year={2023}
}

@inproceedings{zhang2025tiu,
  title={TIU-Bench: A Benchmark for Evaluating Large Multimodal Models on Text-rich Image Understanding},
  author={Zhang, Kun and Niu, Liqiang and Cao, Zhen and Meng, Fandong and Zhou, Jie},
  booktitle={Findings of the Association for Computational Linguistics: EMNLP 2025},
  pages={24286--24295},
  year={2025}
}

@article{huang2025ocr,
  title={Ocr-reasoning benchmark: Unveiling the true capabilities of mllms in complex text-rich image reasoning},
  author={Huang, Mingxin and Shi, Yongxin and Peng, Dezhi and Lai, Songxuan and Xie, Zecheng and Jin, Lianwen},
  journal={arXiv preprint arXiv:2505.17163},
  year={2025}
}

@inproceedings{nguyen2025localizing,
  title={Localizing before answering: A benchmark for grounded medical visual question answering},
  author={Nguyen, Dung and Ho, Minh Khoi and Ta, Huy and Nguyen, Thanh Tam and Chen, Qi and Rav, Kumar and Dang, Quy Duong and Ramchandre, Satwik and Phung, Son Lam and Liao, Zhibin and others},
  booktitle={Thirty-Fourth International Joint Conference on Artificial Intelligence (IJCAI-25)},
  year={2025},
  organization={International Joint Conferences on Artificial Intelligence Organization}
}

@article{qin2025face,
  title={Face-human-bench: A comprehensive benchmark of face and human understanding for multi-modal assistants},
  author={Qin, Lixiong and Ou, Shilong and Zhang, Miaoxuan and Wei, Jiangning and Zhang, Yuhang and Song, Xiaoshuai and Liu, Yuchen and Wang, Mei and Xu, Weiran},
  journal={arXiv preprint arXiv:2501.01243},
  year={2025}
}

@article{zhou2025robotracer,
  title={RoboTracer: Mastering Spatial Trace with Reasoning in Vision-Language Models for Robotics},
  author={Zhou, Enshen and Chi, Cheng and Li, Yibo and An, Jingkun and Zhang, Jiayuan and Rong, Shanyu and Han, Yi and Ji, Yuheng and Liu, Mengzhen and Wang, Pengwei and others},
  journal={arXiv preprint arXiv:2512.13660},
  year={2025}
}

@misc{anywhere3d,
      title={From Objects to Anywhere: A Holistic Benchmark for Multi-level Visual Grounding in 3D Scenes}, 
      author={Tianxu Wang and Zhuofan Zhang and Ziyu Zhu and Yue Fan and Jing Xiong and Pengxiang Li and Xiaojian Ma and Qing Li},
      year={2025},
      eprint={2506.04897},
      archivePrefix={arXiv},
      primaryClass={cs.CV},
      url={https://arxiv.org/abs/2506.04897}, 
}

@inproceedings{ma2025spatialllm,
  title={Spatialllm: A compound 3d-informed design towards spatially-intelligent large multimodal models},
  author={Ma, Wufei and Ye, Luoxin and de Melo, Celso M and Yuille, Alan and Chen, Jieneng},
  booktitle={Proceedings of the Computer Vision and Pattern Recognition Conference},
  pages={17249--17260},
  year={2025}
}

@inproceedings{liu2025visual,
  title={Visual-rft: Visual reinforcement fine-tuning},
  author={Liu, Ziyu and Sun, Zeyi and Zang, Yuhang and Dong, Xiaoyi and Cao, Yuhang and Duan, Haodong and Lin, Dahua and Wang, Jiaqi},
  booktitle={Proceedings of the IEEE/CVF International Conference on Computer Vision},
  pages={2034--2044},
  year={2025}
}

@article{chang2025wearvqa,
  title={WearVQA: A Visual Question Answering Benchmark for Wearables in Egocentric Authentic Real-world scenarios},
  author={Chang, Eun and Huang, Zhuangqun and Liao, Yiwei and Bhavsar, Sagar Ravi and Param, Amogh and Stark, Tammy and Ahmadyan, Adel and Yang, Xiao and Wang, Jiaqi and Abdullah, Ahsan and others},
  journal={arXiv preprint arXiv:2511.22154},
  year={2025}
}

@article{jia2025omnispatial,
  title={Omnispatial: Towards comprehensive spatial reasoning benchmark for vision language models},
  author={Jia, Mengdi and Qi, Zekun and Zhang, Shaochen and Zhang, Wenyao and Yu, Xinqiang and He, Jiawei and Wang, He and Yi, Li},
  journal={arXiv preprint arXiv:2506.03135},
  year={2025}
}

@article{lu2024ovis,
  title={Ovis: Structural embedding alignment for multimodal large language model},
  author={Lu, Shiyin and Li, Yang and Chen, Qing-Guo and Xu, Zhao and Luo, Weihua and Zhang, Kaifu and Ye, Han-Jia},
  journal={arXiv preprint arXiv:2405.20797},
  year={2024}
}

@article{touvron2023llama,
  title={Llama: Open and efficient foundation language models},
  author={Touvron, Hugo and Lavril, Thibaut and Izacard, Gautier and Martinet, Xavier and Lachaux, Marie-Anne and Lacroix, Timoth{\'e}e and Rozi{\`e}re, Baptiste and Goyal, Naman and Hambro, Eric and Azhar, Faisal and others},
  journal={arXiv preprint arXiv:2302.13971},
  year={2023}
}

@article{an2025llava,
  title={Llava-onevision-1.5: Fully open framework for democratized multimodal training},
  author={An, Xiang and Xie, Yin and Yang, Kaicheng and Zhang, Wenkang and Zhao, Xiuwei and Cheng, Zheng and Wang, Yirui and Xu, Songcen and Chen, Changrui and Zhu, Didi and others},
  journal={arXiv preprint arXiv:2509.23661},
  year={2025}
}

@article{wu2024deepseek,
  title={Deepseek-vl2: Mixture-of-experts vision-language models for advanced multimodal understanding},
  author={Wu, Zhiyu and Chen, Xiaokang and Pan, Zizheng and Liu, Xingchao and Liu, Wen and Dai, Damai and Gao, Huazuo and Ma, Yiyang and Wu, Chengyue and Wang, Bingxuan and others},
  journal={arXiv preprint arXiv:2412.10302},
  year={2024}
}

@article{Qwen2.5-VL,
  title={Qwen2.5-VL Technical Report},
  author={Bai, Shuai and Chen, Keqin and Liu, Xuejing and Wang, Jialin and Ge, Wenbin and Song, Sibo and Dang, Kai and Wang, Peng and Wang, Shijie and Tang, Jun and Zhong, Humen and Zhu, Yuanzhi and Yang, Mingkun and Li, Zhaohai and Wan, Jianqiang and Wang, Pengfei and Ding, Wei and Fu, Zheren and Xu, Yiheng and Ye, Jiabo and Zhang, Xi and Xie, Tianbao and Cheng, Zesen and Zhang, Hang and Yang, Zhibo and Xu, Haiyang and Lin, Junyang},
  journal={arXiv preprint arXiv:2502.13923},
  year={2025}
}

@inproceedings{tian2025identifying,
  title={Identifying and mitigating position bias of multi-image vision-language models},
  author={Tian, Xinyu and Zou, Shu and Yang, Zhaoyuan and Zhang, Jing},
  booktitle={Proceedings of the Computer Vision and Pattern Recognition Conference},
  pages={10599--10609},
  year={2025}
}

@article{chaudhary2025investigating,
  title={Investigating Spatial Attention Bias in Vision-Language Models},
  author={Chaudhary, Aryan and Goyal, Sanchit and Narang, Pratik and Kumar, Dhruv},
  journal={arXiv preprint arXiv:2512.18231},
  year={2025}
}

@inproceedings{liu2025dynamic,
  title={Dynamic derivation and elimination: Audio visual segmentation with enhanced audio semantics},
  author={Liu, Chen and Yang, Liying and Li, Peike and Wang, Dadong and Li, Lincheng and Yu, Xin},
  booktitle={Proceedings of the IEEE/CVF Conference on Computer Vision and Pattern Recognition},
  pages={3131--3141},
  year={2025}
}

@inproceedings{liu2025robust,
  title={Robust audio-visual segmentation via audio-guided visual convergent alignment},
  author={Liu, Chen and Li, Peike and Yang, Liying and Wang, Dadong and Li, Lincheng and Yu, Xin},
  booktitle={Proceedings of the Computer Vision and Pattern Recognition Conference},
  pages={28922--28931},
  year={2025}
}

@inproceedings{qiu2024language,
  title={Language-guided multi-modal emotional mimicry intensity estimation},
  author={Qiu, Feng and Zhang, Wei and Liu, Chen and Li, Lincheng and Du, Heming and Guo, Tianchen and Yu, Xin},
  booktitle={Proceedings of the IEEE/CVF Conference on Computer Vision and Pattern Recognition},
  pages={4742--4751},
  year={2024}
}

@inproceedings{liu2024benchmarking,
  title={Benchmarking audio visual segmentation for long-untrimmed videos},
  author={Liu, Chen and Li, Peike Patrick and Yu, Qingtao and Sheng, Hongwei and Wang, Dadong and Li, Lincheng and Yu, Xin},
  booktitle={Proceedings of the IEEE/CVF Conference on Computer Vision and Pattern Recognition},
  pages={22712--22722},
  year={2024}
}

@inproceedings{zhang2024effective,
  title={An effective ensemble learning framework for affective behaviour analysis},
  author={Zhang, Wei and Qiu, Feng and Liu, Chen and Li, Lincheng and Du, Heming and Guo, Tianchen and Yu, Xin},
  booktitle={Proceedings of the IEEE/CVF conference on computer vision and pattern recognition},
  pages={4761--4772},
  year={2024}
}

@inproceedings{guo2026beyond,
  title={Beyond Single-View Sufficiency: CVBench for Cross-View Human Understanding},
  author={Guo, Tianchen and Liu, Chen and Yu, Xin},
  booktitle={Proceedings of the IEEE/CVF Conference on Computer Vision and Pattern Recognition},
  pages={7154--7164},
  year={2026}
}

@inproceedings{liu2024compound,
  title={Compound Expression Recognition via Curriculum Learning},
  author={Liu, Chen and Qiu, Feng and Zhang, Wei and Li, Lincheng and Wang, Dadong and Yu, Xin},
  booktitle={European Conference on Computer Vision},
  pages={282--293},
  year={2024},
  organization={Springer}
}

@inproceedings{liu2024affective,
  title={Affective behaviour analysis via progressive learning},
  author={Liu, Chen and Zhang, Wei and Qiu, Feng and Li, Lincheng and Wang, Dadong and Yu, Xin},
  booktitle={European Conference on Computer Vision},
  pages={366--379},
  year={2024},
  organization={Springer}
}

@article{zhang2024affective,
  title={Affective behaviour analysis via integrating multi-modal knowledge},
  author={Zhang, Wei and Qiu, Feng and Liu, Chen and Li, Lincheng and Du, Heming and Guo, Tiancheng and Yu, Xin},
  journal={arXiv preprint arXiv:2403.10825},
  year={2024}
}

@inproceedings{guo2024being,
  title={Who is being impersonated? Deepfake audio detection and impersonated identification via extraction of id-specific features},
  author={Guo, Tianchen and Du, Heming and Huo, Huan and Liu, Bo and Yu, Xin},
  booktitle={International Conference on Algorithms and Architectures for Parallel Processing},
  pages={301--320},
  year={2024},
  organization={Springer}
}

@inproceedings{qi2026smokebench,
  title={Smokebench: Evaluating multimodal large language models for wildfire smoke detection},
  author={Qi, Tianye and Li, Weihao and Barnes, Nick},
  booktitle={Proceedings of the IEEE/CVF Winter Conference on Applications of Computer Vision},
  pages={1043--1053},
  year={2026}
}

@inproceedings{ke2025dynamic,
  title={Dynamic Orchestration of Multi-agent System for Real-World Multi-image Agricultural VQA},
  author={Ke, Yan and Yu, Xin and Du, Heming and Chapman, Scott and Huang, Helen},
  booktitle={Australasian Database Conference},
  pages={153--165},
  year={2025},
  organization={Springer}
}

@article{hong2026esi,
  title={ESI-Bench: Towards Embodied Spatial Intelligence that Closes the Perception-Action Loop},
  author={Hong, Yining and Liu, Jiageng and Yin, Han and Li, Manling and Guibas, Leonidas and Fei-Fei, Li and Wu, Jiajun and Choi, Yejin},
  journal={arXiv preprint arXiv:2605.18746},
  year={2026}
}

@article{fu2026mme,
  title={Mme: A comprehensive evaluation benchmark for multimodal large language models},
  author={Fu, Chaoyou and Chen, Peixian and Shen, Yunhang and Qin, Yulei and Zhang, Mengdan and Lin, Xu and Yang, Jinrui and Zheng, Xiawu and Li, Ke and Sun, Xing and others},
  journal={Advances in Neural Information Processing Systems},
  volume={38},
  year={2026}
}

@inproceedings{du20253drealcar,
  title={3drealcar: An in-the-wild rgb-d car dataset with 360-degree views},
  author={Du, Xiaobiao and Wang, Yida and Sun, Haiyang and Wu, Zhuojie and Sheng, Hongwei and Wang, Shuyun and Ying, Jiaying and Lu, Ming and Zhu, Tianqing and Zhan, Kun and others},
  booktitle={Proceedings of the IEEE/CVF International Conference on Computer Vision},
  pages={26488--26498},
  year={2025}
}

@article{du2024mvgs,
  title={Mvgs: Multi-view-regulated gaussian splatting for novel view synthesis},
  author={Du, Xiaobiao and Wang, Yida and Yu, Xin},
  year={2024}
}

@article{du2026mobile,
  title={Mobile-GS: Real-time Gaussian splatting for mobile devices},
  author={Du, Xiaobiao and Wang, Yida and Zhan, Kun and Yu, Xin},
  journal={arXiv preprint arXiv:2603.11531},
  year={2026}
}

@article{du2024dreamcar,
  title={Dreamcar: Leveraging car-specific prior for in-the-wild 3d car reconstruction},
  author={Du, Xiaobiao and Sun, Haiyang and Lu, Ming and Zhu, Tianqing and Yu, Xin},
  journal={IEEE Robotics and Automation Letters},
  volume={10},
  number={2},
  pages={1840--1847},
  year={2024},
  publisher={IEEE}
}

@article{du2024ethics,
  title={Ethics-aware face recognition aided by synthetic face images},
  author={Du, Xiaobiao and Yu, Xin and Liu, Jinhui and Dai, Beifen and Xu, Feng},
  journal={Neurocomputing},
  volume={600},
  pages={128129},
  year={2024},
  publisher={Elsevier}
}

@article{chen2024survey,
  title={A survey on 3d gaussian splatting},
  author={Chen, Guikun and Wang, Wenguan},
  journal={arXiv preprint arXiv:2401.03890},
  year={2024}
}

@inproceedings{wu2026metom,
  title={MeToM: Metadata-Guided Token Merging for Efficient Video LLMs},
  author={Wu, Zhuojie and Wang, Shijie and Yu, Xin},
  booktitle={Proceedings of the IEEE/CVF Conference on Computer Vision and Pattern Recognition},
  pages={10441--10450},
  year={2026}
}

@article{wu2022showface,
  title={Showface: Coordinated face inpainting with memory-disentangled refinement networks},
  author={Wu, Zhuojie and Qi, Xingqun and Wang, Zijian and Zhou, Wanting and Yuan, Kun and Sun, Muyi and Sun, Zhenan},
  journal={arXiv preprint arXiv:2204.02824},
  year={2022}
}

@inproceedings{guo2025plnet,
  title={PLNet-12: A Vision-Language Benchmark for Zero-Shot Physical Literacy Analysis Across 12 Fundamental Movements},
  author={Guo, Tianchen and Logan, Peter Anthony and Wackwitz, Thomas and Martin, David},
  booktitle={Australasian Joint Conference on Artificial Intelligence},
  pages={242--254},
  year={2025},
  organization={Springer}
}

@inproceedings{qiu2024learning,
  title={Learning transferable compound expressions from masked autoencoder pretraining},
  author={Qiu, Feng and Du, Heming and Zhang, Wei and Liu, Chen and Li, Lincheng and Guo, Tianchen and Yu, Xin},
  booktitle={Proceedings of the IEEE/CVF Conference on Computer Vision and Pattern Recognition},
  pages={4733--4741},
  year={2024}
}

@article{pan2024large,
  title={A large-scale investigation of semantically incompatible apis behind compatibility issues in android apps},
  author={Pan, Shidong and Guo, Tianchen and Zhang, Lihong and Liu, Pei and Xing, Zhenchang and Sun, Xiaoyu},
  journal={arXiv preprint arXiv:2406.17431},
  year={2024}
}

@article{xu2024m3a,
  title={M3a: A multimodal misinformation dataset for media authenticity analysis},
  author={Xu, Qingzheng and Chen, Huiqiang and Du, Heming and Zhang, Hu and {\L}ukasik, Szymon and Zhu, Tianqing and Yu, Xin},
  journal={Computer Vision and Image Understanding},
  volume={249},
  pages={104205},
  year={2024},
  publisher={Elsevier}
}

@inproceedings{xu2025mdam3,
  title={Mdam3: A misinformation detection and analysis framework for multitype multimodal media},
  author={Xu, Qingzheng and Du, Heming and {\L}ukasik, Szymon and Zhu, Tianqing and Wang, Sen and Yu, Xin},
  booktitle={Proceedings of the ACM on Web Conference 2025},
  pages={5285--5296},
  year={2025}
}

@article{zhang2025mllms,
  title={Why do mllms struggle with spatial understanding? a systematic analysis from data to architecture},
  author={Zhang, Wanyue and Huang, Yibin and Xu, Yangbin and Huang, JingJing and Zhi, Helu and Ren, Shuo and Xu, Wang and Zhang, Jiajun},
  journal={arXiv preprint arXiv:2509.02359},
  year={2025}
}

\clearpage 
\appendix

\begin{center}
    \Large \textbf{Supplementary Material for SSMNBench}
\end{center}


\section{Source Dataset Details}
\label{sec:supp_source_datasets}
SSMNBench (\textbf{S}ingle-view \textbf{S}ufficiency and \textbf{M}ulti-view \textbf{N}ecessity) is built from eight high-quality multi-view datasets to capture the complexity of unstructured real-world environments. These datasets were selected for their complementary strengths in occlusion density, interaction complexity, and viewpoint diversity.

\noindent\textbf{Core4D~\cite{liu2025core4d}.} Core4D is a comprehensive dataset centred on collaborative and unstructured human-object interactions in 3D space. It involves complex real-world scenarios in which multiple individuals manipulate shared objects simultaneously. 
Because these interactions often involve people reaching across one another and occluding the camera view, Core4D is particularly valuable for evaluating cross-view human-object interaction reasoning and relative distance estimation under severe mutual occlusion.

\noindent\textbf{M3GYM~\cite{xu2025m3gym}.} M3GYM is a multi-modal, multi-view dataset focused on high-intensity physical gym exercises. Because athletes frequently perform complex, non-standard body movements (\eg, deep squats, deadlifts, and stretches), the dataset contains severe self-occlusion, with limbs often blocking other anatomical joints.
Its diverse camera angles make M3GYM well-suited for fine-grained action recognition, human counting, and anatomical joint articulation tasks.

\noindent\textbf{Harmony4D~\cite{khirodkar2024harmony4d}.} Harmony4D is a dense, multi-camera video dataset designed to capture nuanced human-human and human-object interactions. 
Using synchronized multi-view camera rigs, it provides high-fidelity spatial observations of individuals in close proximity. 
This makes it well-suited for evaluating MLLMs on close-range physical contact, relative self-pose comparison, and micro-level interaction grounding, helping ensure that benchmark performance reflects geometric alignment rather than semantic inference alone.

\noindent\textbf{Ego-Human~\cite{khirodkar2023ego}.} Ego-Human captures highly dynamic, multi-person scenes across diverse environments. 
While the original dataset includes both wearable and static cameras, our benchmark includes only the synchronized exocentric (third-person static) views.
These wide-baseline exocentric camera arrays provide rich, multi-angle coverage of complex human poses, rapid motions, and intersecting subject trajectories. By focusing entirely on these exocentric views, the dataset allows us to evaluate a model's ability to maintain structural consistency and track high-energy human interactions across disparate spatial angles.

\noindent\textbf{4D-OR~\cite{ozsoy20224d}.} 4D-OR is a pioneering 4D dataset recorded during authentic clinical operations in real operating rooms. The clinical setting naturally creates extremely crowded scenes, uniform clinical clothing (scrubs) that strip away standard texture/color identification cues, and severe mutual occlusion around the operating table. These challenging factors make it a premier source for our action recognition, contact reasoning, distinct human counting and global depth ordering tasks.

\noindent\textbf{MM-OR~\cite{ozsoy2024mmor}.} Building upon the operating room paradigm, MM-OR provides multi-modal sensor data capturing complex, multi-step clinical workflows. This dataset is characterized by dense visual clutter, specialized medical equipment, and highly coordinated team movements. It is critical for testing an MLLM's ability to track subtle, fast-paced human-object interactions, such as passing delicate medical instruments between surgeons, where the object might be partially visible or entirely hidden from a single camera's perspective.

\noindent\textbf{MvMHAT~\cite{gan2021mvmhat}.} The Multi-view Multi-Human Action and Tracking (MvMHAT) dataset provides wide-area spatial coverage of multiple moving subjects across intersecting camera fields of view. The dataset challenges tracking and recognition systems by introducing frequent identity crossovers and background distractors. In SSMNBench, we utilize MvMHAT for testing distinct human counting and global orientation identification, evaluating how well models can track specific entities across disparate visual streams.

\noindent\textbf{HOI-M3~\cite{zhang2024hoi}.} HOI-M3 is a multi-view human-object interaction dataset with fine-grained physical interactions in diverse indoor scenes. 
It is well-suited for evaluating precise human-object contact estimation, especially where interactions are partially obscured by the environment or the subject's body.
Such settings are useful for evaluating whether MLLMs can perceive distance and distinguish genuine contact from cases that appear to involve contact in some views but are non-contact in 3D.

\section{Comparison with Existing Benchmarks}
\label{sec:supp_benchmark_comparison}

To clarify the distinctive role of SSMNBench, Table~\ref{tab:benchmark_comparison} compares its evaluation framework with representative single-image and existing multi-view vision-language benchmarks. 
Most prior benchmarks treat visual inputs as a static set, either through pure 2D perception in single-view settings or through flattened view sequences in multi-view settings. 
Such designs make it difficult to isolate the contribution of individual viewpoints. 
By explicitly annotating \textit{Golden Views} ($\mathcal{V}_{GT}$) and dynamically perturbing view combinations, SSMNBench distinguishes robustness to irrelevant or noisy views (SVS) from the ability to perform genuine cross-view geometric fusion (MVN).

\begin{table}[t]
\centering
\scriptsize
\setlength{\tabcolsep}{3.5pt}
\renewcommand{\arraystretch}{1.0}
\vspace{-1.0em}
\caption{Conceptual comparison of SSMNBench against representative single-image and multi-view MLLM benchmarks. Unlike prior benchmarks, SSMNBench explicitly distinguishes visual sufficiency and necessity through dynamic view perturbation.}
\label{tab:benchmark_comparison}
\begin{tabular}{p{2.8cm}|>{\centering\arraybackslash}p{1.3cm}|>{\centering\arraybackslash}p{1.2cm}|>{\centering\arraybackslash}p{1.3cm}|>{\centering\arraybackslash}p{1.9cm}|>{\centering\arraybackslash}p{1.9cm}}
\toprule
\textbf{Benchmark} & \textbf{Multi-View Input} & \textbf{Human-Centric} & \textbf{Occlusion Focus} & \textbf{Dynamic View Perturbation} & \textbf{SVS/MVN Distinguished} \\
\midrule
MME~\cite{fu2023mme} & $\times$ & Partial & $\times$ & $\times$ & $\times$ \\
OCR-Reasoning~\cite{huang2025ocr} & $\times$ & $\times$ & $\times$ & $\times$ & $\times$ \\
MVBench~\cite{li2024mvbench} & $\checkmark$ & Partial & $\times$ & $\times$ & $\times$ \\
VSI-Bench~\cite{yang2025thinking} & $\checkmark$ & $\times$ & $\times$ & $\times$ & $\times$ \\
Blink~\cite{fu2024blink} & $\checkmark$ & Partial & $\times$ & $\times$ & $\times$ \\
MuirBench~\cite{wang2024muirbench} & $\checkmark$ & Partial & $\times$ & $\times$ & $\times$ \\
Ego3DBench~\cite{gholami2025spatialreasoningvisionlanguagemodels} & $\checkmark$ & $\times$ & $\times$ & $\times$ & $\times$ \\
MMSI-Bench~\cite{yang2025mmsi} & $\checkmark$ & Partial & $\times$ & $\times$ & $\times$ \\
AllAnglesBench~\cite{yeh2025seeing} & $\checkmark$ & Partial & $\times$ & $\times$ & $\times$ \\
\midrule
\textbf{SSMNBench (Ours)} & $\checkmark$ & $\checkmark$ & $\checkmark$ & $\checkmark$ & $\checkmark$ \\
\bottomrule
\end{tabular}
\end{table}
\vspace{-1.0em}

\section{Experimental results with more MLLMs}
\label{sec:supp_extended_results}
As introduced in Section \ref{sec:experiments} of the main manuscript, we present the comprehensive evaluation results for the complete suite of 17 MLLMs. Table~\ref{tab:supp_results_part1} details the performance of the models omitted from the main text due to spatial constraints, specifically highlighting variants from the Qwen~\cite{qwen3.5}, LLaVA~\cite{liu2023llava}, InternVL~\cite{wang2025internvl3_5}, and DeepSeek~\cite{deepseekai2024deepseekv3technicalreport} families. 

Across these extended evaluations, the ``Distraction Degradation'' phenomenon remains a universal architectural bottleneck. Notably, models with smaller parameter counts, such as DeepSeek-Small ($\delta_{dis} = 0.3$) and InternVL3.5-4B ($\delta_{dis} = -0.3$), exhibit a counterintuitive resilience to redundant viewpoints compared to their larger counterparts. However, this robustness is largely symptomatic of a lower baseline accuracy ($\sim$27\%-31\%), suggesting that these models lack the capacity to extract complex cross-attention patterns in the first place, thereby inadvertently ignoring both useful and distracting multi-view cues.

\section{Prompting Details and Evaluation Protocol}
\label{sec:supp_prompts}
To ensure the reproducibility of our quantitative results, we report the exact prompt templates used for querying the MLLMs. In all evaluations, the temperature is fixed to $0.0$ (greedy decoding), ensuring deterministic generation.

\subsection{System Prompt and Task Instruction}
We use a unified zero-shot system prompt and user instruction template across all benchmarked architectures to ensure consistent prompting conditions.

\begin{center}
\begin{minipage}{0.98\linewidth}
\begin{tcolorbox}[
breakable,
title=Prompt Template for SSMNBench Evaluation,
colback=white,
colframe=black!70,
fonttitle=\scriptsize\bfseries
]
{\footnotesize\textbf{System Prompt.}}\vspace{-0.35em}

\begin{tcolorbox}[
colback=black!2,
colframe=black!10,
boxrule=0.25pt,
arc=0.5mm,
left=0.45mm,
right=0.45mm,
top=0.1mm,
bottom=0.1mm
]
{\scriptsize\ttfamily
\setlength{\baselineskip}{8.2pt}
You are an expert in multi-view human and human-object understanding.\\
You must answer strictly with one single letter from [A, B, C, D].\\
Do not add any explanation or extra text. Directly answer.
}
\end{tcolorbox}

{\footnotesize\textbf{Input Template.}}\vspace{-0.35em}

\begin{tcolorbox}[
colback=black!2,
colframe=black!10,
boxrule=0.25pt,
arc=0.5mm,
left=0.45mm,
right=0.45mm,
top=0.1mm,
bottom=0.1mm
]
{\scriptsize\ttfamily
\setlength{\baselineskip}{7.5pt}
You are given synchronized views of the same scene.\\
Question: \{question\}\\
Options: A) \{optA\} \; B) \{optB\} \; C) \{optC\} \; D) \{optD\}\\
Output: Answer with exactly ONE letter from [A, B, C, D], e.g., just "C".\\
Ensure the selected option is consistent with the visual evidence.\\
No explanation. Only one letter.
}
\end{tcolorbox}

\label{fig:prompt_template}
\end{tcolorbox}
\end{minipage}
\end{center}

\begin{table}[htbp]
\centering
\caption{Comprehensive evaluation results on SSMNBench (Supplementary Part 1: Qwen and LLaVA families). Task abbreviations: \textbf{Act.} (Fine-grained Action), \textbf{H-H} (Human-Human Contact), \textbf{H-O} (Human-Object Contact), \textbf{Self} (Relative Self-Pose), \textbf{Joint} (Anatomical Joint), \textbf{Count} (Distinct Human Counting), \textbf{Attr.} (Counting with Attribute), \textbf{Cross} (Relative Cross-Person Pose), \textbf{Dist.} (Relative Distance), \textbf{Ori.} (Global Orientation), \textbf{Depth} (Global Depth Ordering). ``--'' indicates the setting is not applicable. $\delta_{dis} \downarrow$ represents the overall Distraction Decay (lower is better).}
\label{tab:supp_results_part1}
\resizebox{\textwidth}{!}{
\begin{tabular}{@{}ll ccccc|c cccccc|c | c@{}}
\toprule
\multirow{2}{*}{\textbf{Model}} & \multirow{2}{*}{\textbf{Setting}} & \multicolumn{6}{c}{\textbf{SVS Tasks}} & \multicolumn{7}{c}{\textbf{MVN Tasks}} & \multirow{2}{*}{$\delta_{dis} \downarrow$} \\
\cmidrule(lr){3-8} \cmidrule(lr){9-15}
& & Act. & H-H & H-O & Self & Joint & \textbf{Avg} & Count & Attr. & Cross & Dist. & Ori. & Depth & \textbf{Avg} & \\
\midrule

\multirow{5}{*}{Qwen3-4B} 
& -1 (MVN only) & -- & -- & -- & -- & -- & -- & 25.7 & 33.7 & 36.0 & 54.0 & 28.7 & 31.7 & 35.0 & \multirow{5}{*}{1.4} \\
& +0 (GT views) & 36.3 & 36.7 & 36.7 & 38.7 & 36.7 & 37.0 & 34.3 & 35.3 & 38.0 & 45.7 & 24.3 & 33.0 & 35.1 & \\
& +1 view       & 36.7 & 32.3 & 36.7 & 35.7 & 37.3 & 35.7 & 34.0 & 32.3 & 37.0 & 44.7 & 24.0 & 33.7 & 34.3 & \\
& +2 views      & 35.7 & 28.7 & 38.0 & 35.7 & 35.3 & 34.7 & 37.3 & 33.3 & 37.7 & 40.7 & 23.3 & 35.0 & 34.6 & \\
& +3 (SVS only) & 33.3 & 30.3 & 36.7 & 35.3 & 35.3 & 34.2 & -- & -- & -- & -- & -- & -- & -- & \\
\midrule

\multirow{5}{*}{Qwen3-8B} 
& -1 (MVN only) & -- & -- & -- & -- & -- & -- & 32.7 & 28.7 & 33.7 & 51.7 & 29.0 & 35.7 & 35.2 & \multirow{5}{*}{2.2} \\
& +0 (GT views) & 35.7 & 30.7 & 41.3 & 38.7 & 36.3 & 36.5 & 37.7 & 34.7 & 36.7 & 47.7 & 24.7 & 37.0 & 36.4 & \\
& +1 view       & 31.3 & 28.7 & 39.3 & 35.3 & 36.3 & 34.2 & 37.3 & 34.7 & 36.7 & 44.7 & 24.7 & 32.0 & 35.0 & \\
& +2 views      & 32.0 & 26.7 & 38.7 & 36.3 & 34.7 & 33.7 & 37.7 & 34.7 & 34.3 & 43.7 & 24.0 & 31.7 & 34.4 & \\
& +3 (SVS only) & 31.3 & 26.0 & 39.7 & 35.3 & 33.7 & 33.2 & -- & -- & -- & -- & -- & -- & -- & \\
\midrule

\multirow{5}{*}{Qwen2.5-32B} 
& -1 (MVN only) & -- & -- & -- & -- & -- & -- & 36.0 & 33.3 & 42.0 & 42.0 & 25.7 & 39.0 & 36.3 & \multirow{5}{*}{2.7} \\
& +0 (GT views) & 42.0 & 34.7 & 48.0 & 38.7 & 38.3 & 40.3 & 44.3 & 38.0 & 44.7 & 39.7 & 26.0 & 40.7 & 38.9 & \\
& +1 view       & 40.7 & 32.3 & 44.3 & 38.7 & 37.7 & 38.7 & 37.7 & 37.3 & 41.0 & 43.0 & 23.0 & 39.0 & 36.8 & \\
& +2 views      & 38.7 & 30.7 & 42.7 & 37.3 & 36.0 & 37.1 & 37.3 & 36.7 & 37.7 & 35.0 & 25.3 & 40.0 & 35.3 & \\
& +3 (SVS only) & 38.0 & 30.3 & 44.0 & 37.3 & 37.0 & 37.3 & -- & -- & -- & -- & -- & -- & -- & \\
\midrule

\multirow{5}{*}{LLaVA-7B} 
& -1 (MVN only) & -- & -- & -- & -- & -- & -- & 31.0 & 31.7 & 30.7 & 44.7 & 18.3 & 25.0 & 30.2 & \multirow{5}{*}{1.3} \\
& +0 (GT views) & 31.7 & 32.0 & 34.7 & 30.3 & 30.7 & 31.9 & 35.3 & 33.7 & 28.0 & 38.0 & 19.0 & 25.0 & 29.8 & \\
& +1 view       & 31.7 & 32.3 & 30.7 & 28.3 & 31.3 & 30.9 & 29.7 & 35.0 & 30.7 & 34.0 & 19.7 & 26.0 & 29.2 & \\
& +2 views      & 29.3 & 32.3 & 32.0 & 29.7 & 30.0 & 30.7 & 28.3 & 36.0 & 28.3 & 31.7 & 19.0 & 22.7 & 27.7 & \\
& +3 (SVS only) & 25.3 & 34.7 & 33.3 & 30.7 & 28.7 & 30.5 & -- & -- & -- & -- & -- & -- & -- & \\
\midrule

\multirow{5}{*}{InternVL3-78B} 
& -1 (MVN only) & -- & -- & -- & -- & -- & -- & 28.0 & 30.3 & 37.0 & 51.0 & 31.0 & 42.0 & 36.6 & \multirow{5}{*}{1.7} \\
& +0 (GT views) & 34.0 & 30.3 & 38.7 & 40.3 & 37.7 & 36.2 & 46.0 & 40.7 & 39.0 & 44.0 & 27.3 & 38.7 & 39.3 & \\
& +1 view       & 33.7 & 28.7 & 38.0 & 38.7 & 35.3 & 34.9 & 42.3 & 41.0 & 39.0 & 45.0 & 27.3 & 36.0 & 38.4 & \\
& +2 views      & 32.0 & 28.0 & 37.7 & 39.7 & 34.3 & 34.3 & 40.3 & 40.3 & 39.7 & 41.7 & 26.0 & 33.7 & 36.9 & \\
& +3 (SVS only) & 31.3 & 27.7 & 37.3 & 38.0 & 35.7 & 34.0 & -- & -- & -- & -- & -- & -- & -- & \\
\midrule

\multirow{5}{*}{InternVL35-4B} 
& -1 (MVN only) & -- & -- & -- & -- & -- & -- & 30.7 & 28.0 & 30.7 & 45.7 & 23.3 & 32.0 & 31.7 & \multirow{5}{*}{-0.3} \\
& +0 (GT views) & 31.7 & 29.3 & 30.3 & 32.0 & 29.7 & 30.6 & 32.3 & 30.7 & 30.0 & 41.0 & 23.0 & 30.7 & 31.3 & \\
& +1 view       & 34.0 & 31.7 & 30.3 & 33.7 & 30.0 & 31.9 & 28.7 & 28.7 & 31.7 & 41.7 & 22.3 & 34.0 & 31.2 & \\
& +2 views      & 30.7 & 32.3 & 31.7 & 34.0 & 30.7 & 31.9 & 28.3 & 25.7 & 36.7 & 38.0 & 23.0 & 34.0 & 30.9 & \\
& +3 (SVS only) & 31.0 & 28.0 & 29.7 & 33.3 & 29.3 & 30.3 & -- & -- & -- & -- & -- & -- & -- & \\
\midrule

\multirow{5}{*}{InternVL35-14B} 
& -1 (MVN only) & -- & -- & -- & -- & -- & -- & 28.0 & 26.3 & 34.3 & 55.0 & 28.7 & 32.7 & 34.2 & \multirow{5}{*}{1.9} \\
& +0 (GT views) & 34.7 & 22.7 & 38.7 & 36.3 & 32.7 & 33.0 & 33.7 & 32.3 & 33.0 & 43.0 & 26.7 & 32.0 & 33.4 & \\
& +1 view       & 34.0 & 22.0 & 39.3 & 36.3 & 32.3 & 32.8 & 27.7 & 31.3 & 28.0 & 42.0 & 23.0 & 33.0 & 30.8 & \\
& +2 views      & 32.3 & 22.3 & 36.3 & 35.3 & 32.7 & 31.8 & 27.3 & 30.3 & 28.3 & 39.7 & 23.3 & 34.0 & 30.5 & \\
& +3 (SVS only) & 29.7 & 22.0 & 37.3 & 37.3 & 31.3 & 31.5 & -- & -- & -- & -- & -- & -- & -- & \\
\midrule

\multirow{5}{*}{DeepSeek-Tiny} 
& -1 (MVN only) & -- & -- & -- & -- & -- & -- & 17.0 & 30.0 & 33.0 & 37.7 & 22.7 & 26.7 & 27.8 & \multirow{5}{*}{0.5} \\
& +0 (GT views) & 33.7 & 32.7 & 31.3 & 25.3 & 25.0 & 29.6 & 19.3 & 27.0 & 33.0 & 35.7 & 24.3 & 26.0 & 27.6 & \\
& +1 view       & 34.7 & 29.7 & 29.7 & 28.7 & 25.7 & 29.7 & 19.3 & 26.3 & 31.7 & 35.0 & 25.7 & 25.7 & 27.3 & \\
& +2 views      & 33.0 & 32.0 & 32.7 & 26.3 & 25.3 & 29.9 & 18.3 & 23.3 & 31.0 & 35.0 & 24.0 & 24.0 & 25.9 & \\
& +3 (SVS only) & 32.7 & 30.0 & 31.3 & 25.7 & 25.3 & 29.0 & -- & -- & -- & -- & -- & -- & -- & \\
\midrule

\multirow{5}{*}{DeepSeek-Small} 
& -1 (MVN only) & -- & -- & -- & -- & -- & -- & 26.7 & 28.0 & 32.7 & 40.0 & 26.0 & 26.7 & 30.0 & \multirow{5}{*}{0.3} \\
& +0 (GT views) & 30.0 & 18.3 & 32.0 & 24.3 & 33.3 & 27.6 & 27.0 & 30.0 & 30.7 & 39.0 & 21.7 & 27.7 & 29.3 & \\
& +1 view       & 34.3 & 19.7 & 30.3 & 25.7 & 32.3 & 28.5 & 26.7 & 28.0 & 31.0 & 37.0 & 22.7 & 29.0 & 29.1 & \\
& +2 views      & 28.0 & 19.7 & 29.3 & 23.7 & 33.7 & 26.9 & 28.3 & 27.3 & 29.7 & 37.7 & 21.0 & 29.0 & 28.8 & \\
& +3 (SVS only) & 28.3 & 19.7 & 28.7 & 24.0 & 33.7 & 26.9 & -- & -- & -- & -- & -- & -- & -- & \\
\bottomrule

\end{tabular}
}
\end{table}



\subsection{LLM-Based Fallback Extraction and Statistics}
\label{sec:supp_fallback}
Despite the explicit constraints in the zero-shot system prompts, certain MLLMs occasionally fail to output a strictly compliant single-letter response, instead generating verbose explanations. When the primary rule-based parser fails to extract an exact match, we utilize a lightweight LLM fallback strategy (powered by Gemini-2.5-Flash-Lite) to interpret the raw text output and extract the intended multiple-choice letter.



\begin{center}
\begin{minipage}{1.0\linewidth}
\begin{tcolorbox}[
breakable,
title=Fallback Prompt for Answer Extraction,
colback=white,
colframe=black!70,
fonttitle=\scriptsize\bfseries
]


\begin{tcolorbox}[
colback=black!2,
colframe=black!10,
boxrule=0.25pt,
arc=0.5mm,
left=0.45mm,
right=0.45mm,
top=0.1mm,
bottom=0.1mm
]
{\scriptsize\ttfamily
\setlength{\baselineskip}{8.0pt}
You are an answer extractor. Your task is to identify the final selected option from the provided model output.\\
If the prediction explicitly contains A, B, C, or D, return the corresponding letter.\\
If the prediction does not specify a choice or indicates that none of the options are correct, return None.\\
Output ONLY the single letter (A, B, C, or D) or None. Do not provide explanations.
}
\end{tcolorbox}

\label{fig:fallback_prompt}
\end{tcolorbox}
\end{minipage}
\end{center}

\begin{table}[t]
\centering
\scriptsize
\setlength{\tabcolsep}{5.5pt}
\renewcommand{\arraystretch}{1.2}
\caption{LLM-based fallback extraction trigger ratio (\%). A higher value indicates weaker adherence to the strict single-character output constraint.}
\label{tab:fallback_stats}
\begin{tabular}{lr!{\color{gray!60}\vrule width 0.6pt\hspace{0.4em}}lr}
\toprule
\textbf{Model} & \textbf{Trigger Ratio (\%)} & \textbf{Model} & \textbf{Trigger Ratio (\%)} \\
\midrule
InternVL35-14B   & 0.00 & Qwen2.5-7B      & 0.00 \\
InternVL35-4B    & 0.00  & Qwen2.5-72B     & 0.00 \\
Gemini-2.5-Flash & 0.08  & InternVL3.5-78B & 0.00 \\
InternVL3.5-38B  & 0.00  & Qwen3-4B        & 0.00 \\
Gemini-2.5-Pro   & 0.00  & Qwen3-8B        & 0.00 \\
GPT-5.2          & 0.00  & Qwen2.5-32B     & 0.00 \\
Qwen3-32B        & 0.00  & LLaVA-7B        & 0.00 \\
InternVL3-78B    & 0.00  & DeepSeek-Tiny   & 0.00 \\
DeepSeek-Small   & 0.00  &                 &      \\
\bottomrule
\end{tabular}
\end{table}

Table~\ref{tab:fallback_stats} reports the fallback trigger rates (\ie, the frequency of employing an auxiliary LLM for answer extraction) across the evaluated models. Notably, both proprietary and open-source models demonstrate near-perfect instruction adherence; with the exception of a negligible 0.08\% rate for Gemini-2.5-Flash, all models required zero fallback interventions (0.00\%).

\section{Detailed Failure Case Analysis}
\label{sec:supp_error_analysis}
To provide a deeper understanding of the specific geometric and semantic bottlenecks hindering current MLLMs, we expand upon the manual error audit discussed in Section~\ref{sec:error_analysis} of the main text. By categorizing failures into \textit{Image-Level} and \textit{View-Level} dimensions, we expose exactly how architectures falter when processing unstructured cross-view images.

\subsubsection{Image-level Errors} reflect fundamental perception failures that arise before, or independently of, cross-view integration. We identify five primary bottlenecks:
\begin{itemize}
    \item \textbf{Spatial Analysis Failure (33.1\%):} The most prevalent source of error. Models consistently fail to infer accurate 3D spatial relationships from 2D projections, struggling with relative height estimation, directional awareness (e.g., left vs. right disambiguation), and monocular depth perception.
    \item \textbf{Contact Area Grounding (21.1\%):} Models frequently fail to identify and localize precise physical contact areas. They struggle to ground the physical boundaries between humans and objects, often hallucinating interactions or missing subtle physical connections entirely.
    \item \textbf{Fine-Grained Human Joint Recognition (19.3\%):} MLLMs show brittleness in micro-level pose understanding. Errors here involve misidentifying specific anatomical joints, misjudging joint status (e.g., flexed vs. extended), or incorrectly naming the localized joint region.
    \item \textbf{Cross-Image Entity Linking (14.7\%):} Models lack the instance-level consistency required to track entities across coordinate spaces. They frequently fail to link the corresponding human, object, or joint across different images, conflating distinct objects or duplicating identical ones.
    \item \textbf{Partial Visibility \& Appearance (12.8\%):} A persistent challenge in occlusion-heavy scenes. Models struggle to accurately recognize human appearance or identity when subjects are only partially visible or heavily truncated by the camera framing.
\end{itemize}

\subsubsection{View-Level Errors}
View-level errors expose the brittleness of current multi-image attention mechanisms when tasked with synthesizing collaborative geometric evidence. 
\begin{itemize}
    \item \textbf{Conflict Information Awareness \& Fusion Failure (67.2\%):} This is the dominant systemic failure in multi-view reasoning. While models often demonstrate an awareness of conflicting visual information across different viewpoints, they lack the geometric inductive biases to resolve it. Consequently, they fail to merge the fragmented cues and are unable to identify or trust the clearest, most informative viewpoint, leading to indecision or hallucination.
    \item \textbf{Over-Reliance on Preferred View (32.8\%):} Rather than synthesizing all available evidence, models frequently exhibit a severe positional or semantic bias toward a single ``preferred'' view. The model extracts an initial observation from this anchor view and forcibly applies it to the others, generating textual responses that seemingly ignore the contradictory or complementary visual evidence actually present in the secondary frames.
\end{itemize}

\section{SSMNBench License}
\label{sec:supp_licenses}
SSMNBench is released strictly as a research benchmark intended for non-commercial, academic use. The diverse human-centric scenes are sourced from open-source multi-view datasets, including Core4D, M3GYM, Harmony4D, Ego-Human, 4D-OR, MM-OR, MvMHAT, and HOI-M3. We have rigorously reviewed and signed the respective data use agreements for all underlying source data. 


\section{More Details about SSMNBench}
\subsection{Diversity}
SSMNBench explicitly targets complex, occlusion-heavy human-centric scenarios. By curating from eight diverse multi-view datasets, we ensure broad spatial coverage across indoor (72.1\%) and outdoor (27.9\%) settings with high complexity (avg. 5.8 persons/image, 47.1\% occlusion). Our selection yields a diverse taxonomy of approximately 438 actions, 242 objects, and 291 subjects. The proposed benchmark proves existing multi-view data is already highly challenging: leading MLLMs only achieve 44.9\% accuracy, which is 48.5\% behind human performance. Therefore, new evaluation protocols (SVS vs. MVN) on collected existing data effectively expose the limitations of current models, given that the benchmark currently presents significant headroom. We will incorporate new data when the performance saturates in the future.

\subsection{Task Construction Details}
The ``golden views'' in the SVS and MVN tasks are randomly selected, while ensuring that humans can answer the questions based on the selected views. This design avoids bias toward any specific viewpoints. Moreover, the additional views in MVN are also randomly selected without introducing view-specific bias. For efficiency and simplicity, the human performance is conducted on a subset.

\subsection{Validity of Multiple-Choice Formulation}
We adopt the multiple-choice format to ensure fair, objective, and reproducible quantitative comparisons across 17 MLLMs, following prior benchmarks~\cite{yeh2025seeing,guo2026beyond,wang2024muirbench,fu2024blink,fu2023mme,yang2025mmsi,li2024mvbench,jia2025omnispatial}. 
We further design a distractor generation strategy to make the answer options sufficiently challenging. 
The validity of the multiple-choice formulation is supported by the substantial gap to human performance, where the best-performing model still lags behind humans by 48.5 percentage points. 
This suggests that our multiple-choice design does not trivialize the task and provides a reasonable protocol for fair, objective, and reproducible evaluation.

\subsection{API Versions of Proprietary Models}
In this paper's experiment section, the exact API versions of Gemini and GPT models are \textit{gemini-2.5-pro}, \textit{gemini-2.5-flash} and \textit{gpt-5.2-2025-12-11}.

\subsection{Annotation Details}
All six annotators completed an 8-hour training on 220 pilot samples. Each QA pair and its corresponding ground-truth view set ($\mathcal{V}_{GT}$) is independently reviewed by three additional annotators. 
Disagreements are resolved via majority voting.
Cases without a clear consensus are discarded.
During the verification phase, approximately 32.7\% of initial samples are revised for clarity, while 13.1\% are entirely discarded because text-solvable or ambiguous.


\subsection{Annotation Interface}
\begin{figure}
  \makebox[\textwidth][c]{\includegraphics[height=5.2cm]{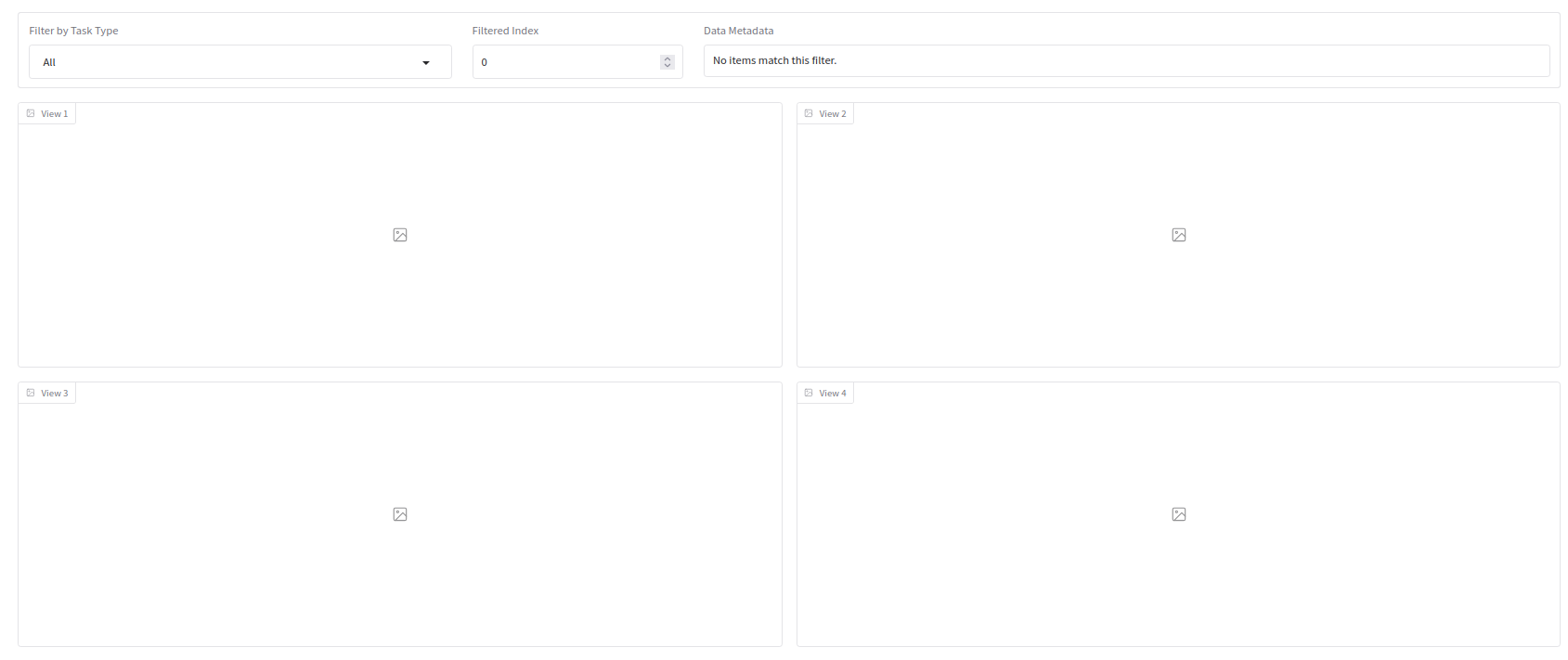}}
  \caption{Annotation Interface (Part 1).}
  \label{fig:interface1}
\end{figure}

\begin{figure}
  \makebox[\textwidth][c]{\includegraphics[height=4.1cm]{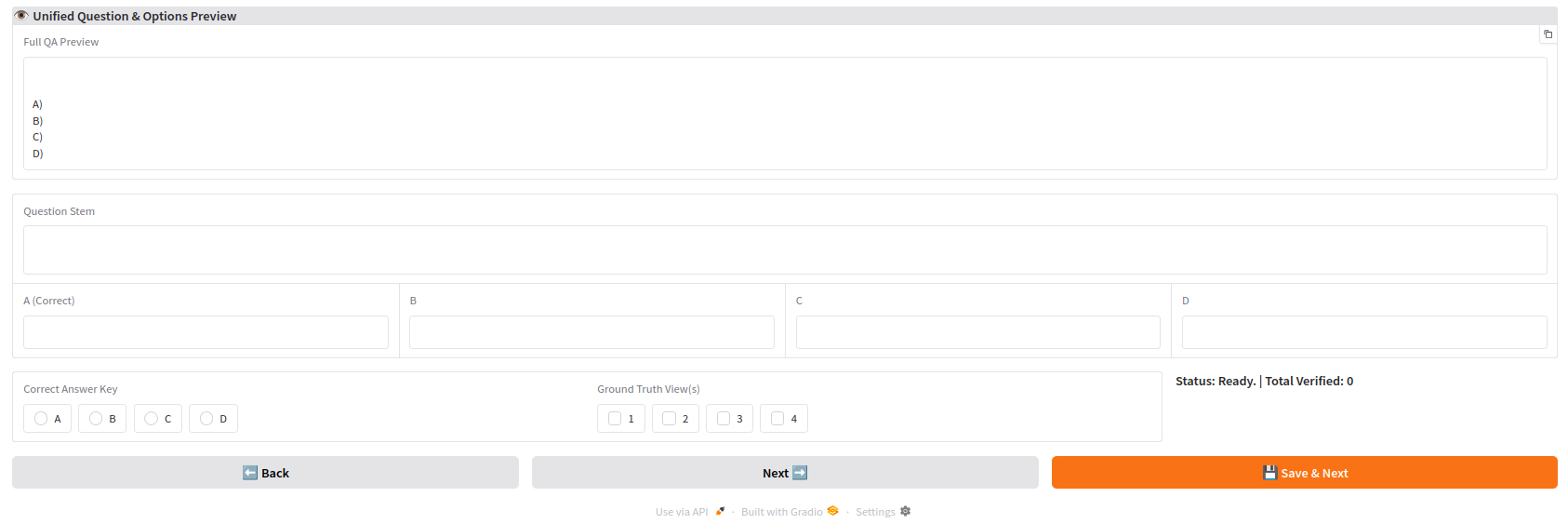}}
  \caption{Annotation Interface (Part 2).}
  \label{fig:interface2}
\end{figure}

The annotation interface is shown in Figure~\ref{fig:interface1} and Figure~\ref{fig:interface2}.

\subsection{Dataset Annotation Format}
\begin{figure}[htbp]
  \centering
  \makebox[\textwidth][c]{\includegraphics[height=8.4cm]{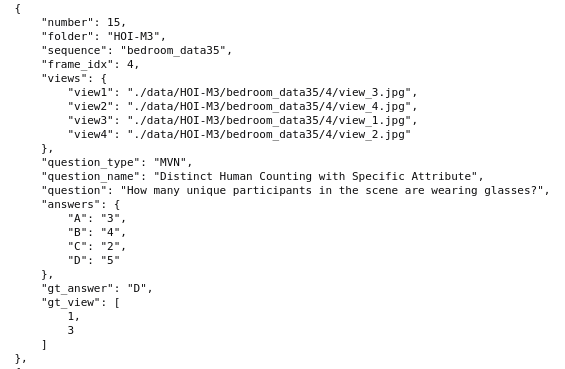}}
  \caption{An illustrative example of the JSON annotation structure utilized in the benchmark. Each entry encapsulates the scene metadata, the natural language query, four multiple-choice options, the ground-truth answer, and the specific camera views required for inference.}
  \label{fig:json}
\end{figure}

The finalized benchmark is structured and serialized in the JavaScript Object Notation (JSON) format to facilitate standardized parsing, interoperability, and automated evaluation. A representative example of a single annotated instance, demonstrating the key-value pairings for the multi-view metadata, question text, candidate options, and ground-truth answer, is illustrated in Figure~\ref{fig:json}.

\section{Additional Benchmark Examples}
\label{sec: additional examples}

\begin{figure}[p]
  \makebox[\textwidth][c]{\includegraphics[height=12.0cm]{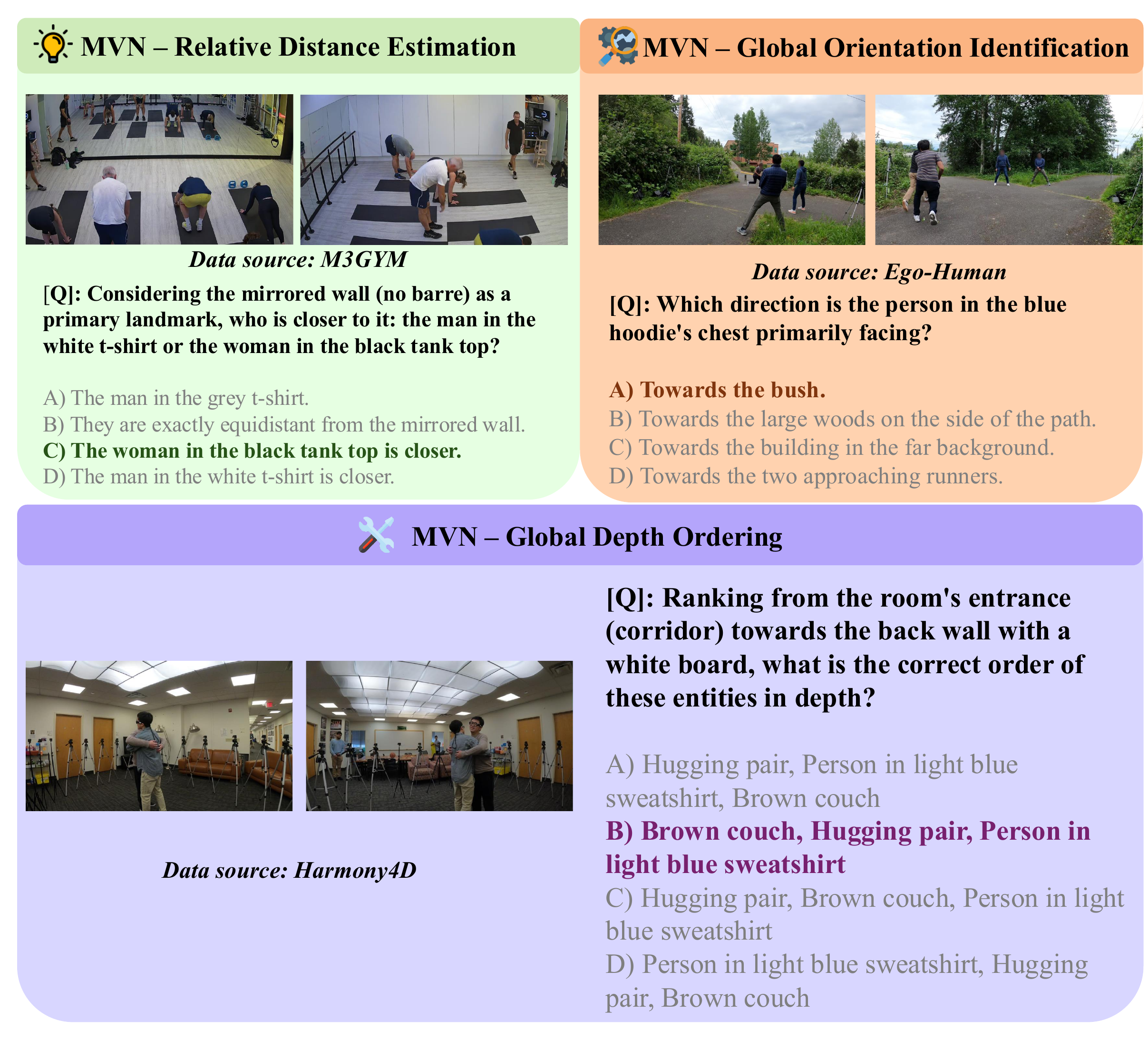}}
  \vspace{-0.5em}
  \caption{SSMNBench Examples (Part 3).}
  \label{fig:91011}
\end{figure}

Further examples of benchmark visualizations are provided below in Figure~\ref{fig:1234}, Figure~\ref{fig:5678}, and Figure~\ref{fig:91011}.

\begin{figure}
  \makebox[\textwidth][c]{\includegraphics[height=12.0cm]{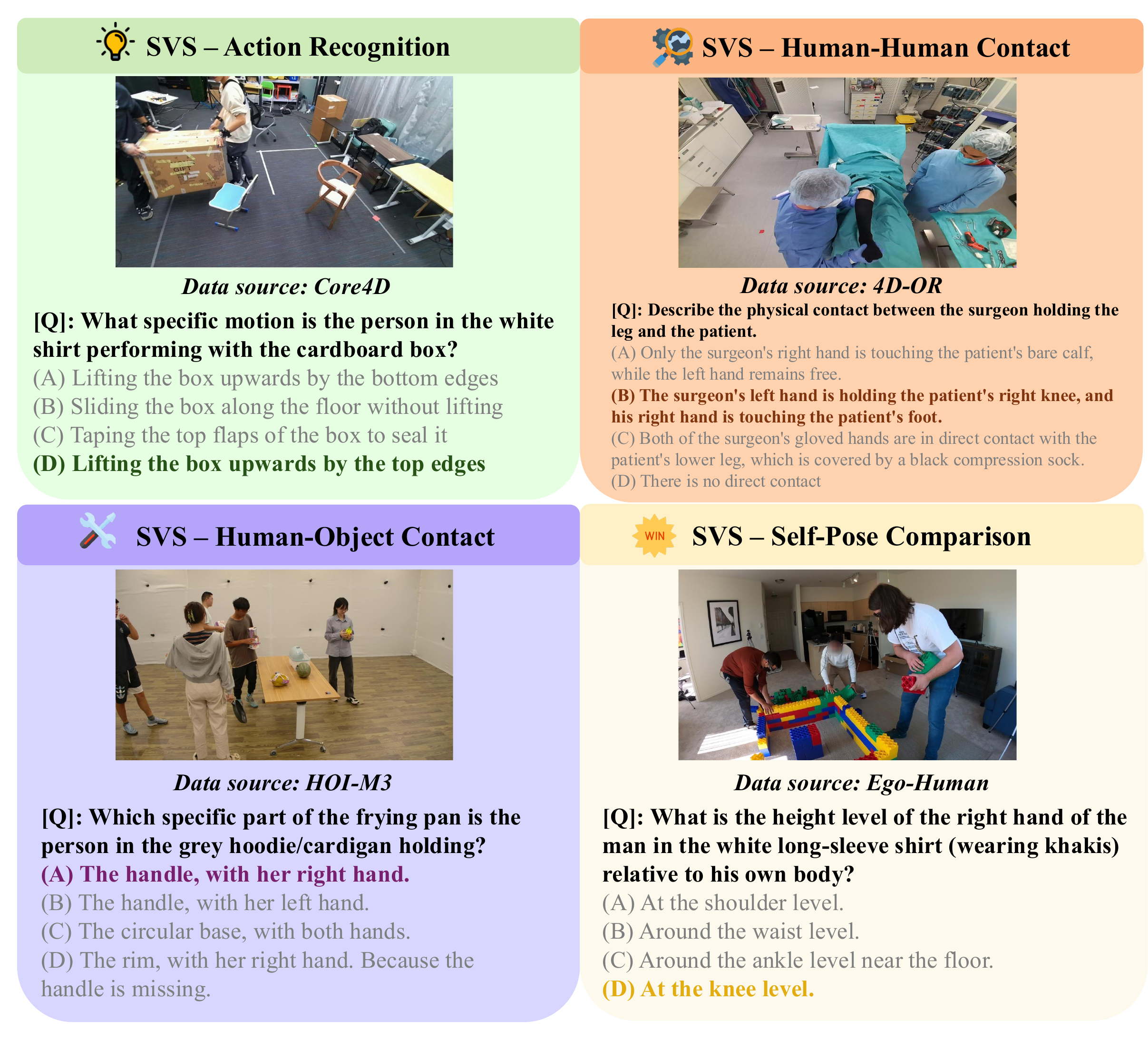}}
  \vspace{-0.5em}
  \caption{SSMNBench Examples (Part 1).}
  \label{fig:1234}
\end{figure}

\begin{figure}[p]
  \makebox[\textwidth][c]{\includegraphics[height=12.0cm]{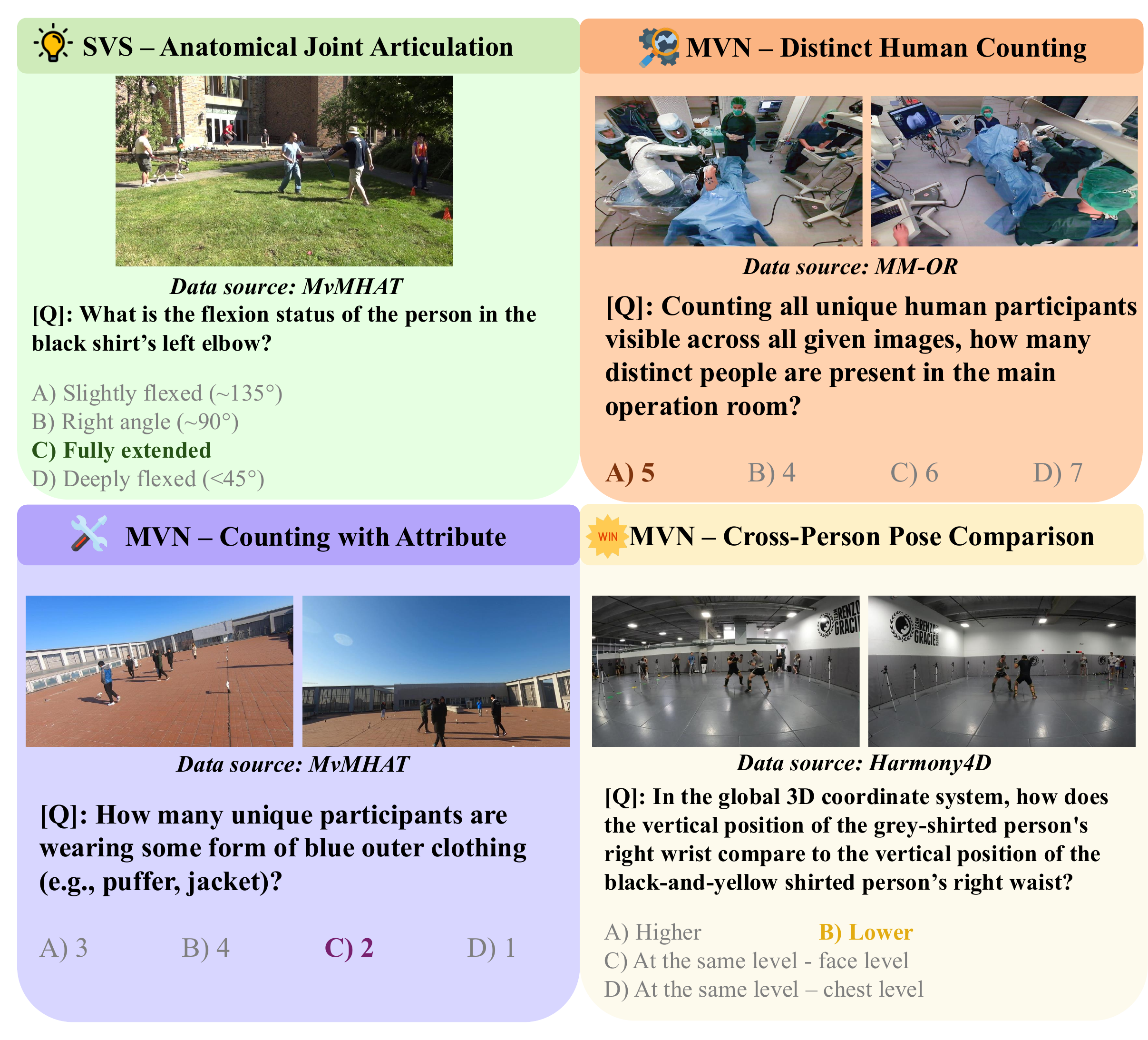}}
  \vspace{-0.5em}
  \caption{SSMNBench Examples (Part 2).}
  \label{fig:5678}
\end{figure}

\end{document}